\PassOptionsToPackage{table}{xcolor}

\documentclass[a4paper,fleqn]{cas-dc}

\usepackage[numbers,sort&compress]{natbib}
\usepackage{algorithm}
\usepackage{algorithmic}
\usepackage{amsmath}
\usepackage{amssymb}
\usepackage{bm}                
\usepackage{graphicx}          
\usepackage{caption}           
\usepackage{subcaption}        
\usepackage[section]{placeins} 
\usepackage{float}             
\usepackage{enumitem}          
\usepackage{longtable}         
\usepackage{xcolor}            
\usepackage{array}            

\begin{document}
\let\WriteBookmarks\relax
\def\floatpagepagefraction{1}
\def\textpagefraction{.001}

\renewcommand{\topfraction}{0.9}       
\renewcommand{\bottomfraction}{0.8}    
\renewcommand{\textfraction}{0.07}     
\renewcommand{\floatpagefraction}{0.7} 
\setcounter{topnumber}{9}              
\setcounter{bottomnumber}{9}           
\setcounter{totalnumber}{20}           
\setcounter{dbltopnumber}{9}           

\shorttitle{Cross-Platform LF-PID Tuning via Meta-RL}

\shortauthors{Wu et al.}

    \title[mode=title]{Cross-Platform Learnable Fuzzy Gain-Scheduled Proportional-Integral-Derivative Controller Tuning via Physics-Constrained Meta-Learning and Reinforcement Learning Adaptation}

\author[1]{Jiahao Wu}[orcid=0009-0003-0555-2461]
\cormark[1]
\ead{13620926353@163.com}
\credit{Conceptualization, Methodology, Software, Validation, Writing - Original Draft, Visualization}

\affiliation[1]{organization={The University of Hong Kong},
            city={Hong Kong},
            country={China}}

\author[2]{Shengwen Yu}
\ead{13823343109@163.com}
\credit{Validation, Writing - Review \& Editing, Data Curation}

\affiliation[2]{organization={Guangzhou College of Commerce},
            city={Guangzhou},
            state={Guangdong},
            country={China}}

\cortext[1]{Corresponding author}

\begin{abstract}
    	\textbf{Motivation and gap:} PID-family controllers remain a pragmatic choice for many robotic systems due to their simplicity and interpretability, but tuning stable, high-performing gains is time-consuming and typically non-transferable across robot morphologies, payloads, and deployment conditions. Fuzzy gain scheduling can provide interpretable online adjustment, yet its per-joint scaling and consequent parameters are platform-dependent and difficult to tune systematically.

        	\textbf{Proposed approach:} We propose a hierarchical framework for cross-platform tuning of a \emph{learnable fuzzy gain-scheduled PID} (LF-PID). The controller uses shared fuzzy membership partitions to preserve common error semantics, while learning per-joint scaling and Takagi--Sugeno consequent parameters that schedule PID gains online. Combined with physics-constrained virtual robot synthesis, meta-learning provides cross-platform initialization from robot physical features, and a lightweight reinforcement learning (RL) stage performs deployment-specific refinement under dynamics mismatch. Starting from three base simulated platforms, we generate 232 physically valid training variants via bounded perturbations of mass (±10\%), inertia (±15\%), and friction (±20\%).

        	\textbf{Results and insight:} We evaluate cross-platform generalization on two distinct systems (a 9-DOF serial manipulator and a 12-DOF quadruped) under multiple disturbance scenarios. The RL adaptation stage improves tracking performance on top of the meta-initialized controller, with up to 80.4\% error reduction in challenging high-load joints (12.36°→2.42°) and 19.2\% improvement under parameter uncertainty. We further identify an \textit{optimization ceiling effect}: online refinement yields substantial gains when the meta-initialized baseline exhibits localized deficiencies, but provides limited improvement when baseline quality is already uniformly strong.
\end{abstract}

\begin{highlights}
\item \textbf{Physics-constrained virtual synthesis:} Generates 232 physically valid robot variants from 3 base platforms via bounded parameter perturbations
\item \textbf{Learnable fuzzy PID design:} Uses shared fuzzy membership partitions with per-joint scaling and consequent parameters for gain scheduling
\item \textbf{Hierarchical meta-RL tuning:} Meta-learns cross-platform LF-PID initialization and refines it via online RL adaptation
\item \textbf{Cross-platform evaluation:} Validated on heterogeneous robots (9-DOF manipulator and 12-DOF quadruped) under multiple disturbances
\item \textbf{Robustness gains:} Achieves up to 80.4\% error reduction on challenging joints and 19.2\% improvement under parameter uncertainty
\item \textbf{Design insight:} Identifies an \textit{optimization ceiling effect} characterizing when RL refinement is most beneficial
\end{highlights}


\maketitle

\section{Introduction}
\label{sec:intro}

\subsection{Motivation: Cross-Platform Tuning Under Dynamics Mismatch}

PID controllers remain widely used in robotics due to their simplicity, interpretability, and ease of deployment \cite{astrom2006advanced}. However, tuning stable and high-performing gains for a new robot is often time-consuming and must be repeated whenever morphology, payload, or contact conditions change \cite{vilanova2012pid,johnson2021industrial}. This lack of transferability becomes more pronounced as robots diversify in kinematics and actuation.

Fuzzy gain-scheduled PID variants can improve robustness by adapting gains online using interpretable rule-based structure, but their per-joint scaling and consequent parameters still require substantial tuning and are often platform-specific.

Classical tuning heuristics and optimization-based auto-tuning \cite{gaing2004particle,berkenkamp2016safe} can reduce manual effort but typically require per-platform re-execution and do not directly address deployment-time model mismatch. Learning-based control can adapt online, yet pure reinforcement learning commonly needs large interaction budgets \cite{lillicrap2015continuous}. Meta-learning can improve reuse across tasks, but it depends on diverse training tasks and may generate unrealistic training distributions without physically grounded augmentation. These limitations motivate an approach that (i) preserves physical plausibility during training data generation, (ii) provides cross-platform initialization, and (iii) supports deployment-time refinement under disturbances.

\subsection{Proposed Solution and Key Contributions}

This paper presents a hierarchical meta-reinforcement learning framework built on \textit{physics-constrained virtual robot synthesis}. By generating 232 dynamically-valid robot variants from only three simulated base platforms, we obtain a physically grounded training distribution for data-efficient meta-learning, while maintaining plausibility through bounded parameter perturbations.

\textbf{Key technical contributions:}

\begin{enumerate}[leftmargin=*, itemsep=2pt]
    \item \textbf{Physics-based virtual synthesis} generating 232 training variants from 3 base robots via constrained perturbation (mass ±10\%, inertia ±15\%, friction ±20\%), enabling physically grounded and data-efficient meta-learning
    \item \textbf{Hierarchical two-stage architecture} decoupling cross-platform initialization (meta-learning) from deployment-specific refinement (RL), achieving 80.4\% error reduction in challenging high-load joints
    \item \textbf{Optimization ceiling effect discovery} revealing RL provides significant gains (16.6\%) for localized high-error joints but minimal benefit (2.1\%) for uniformly strong baselines—establishing practical design guidance for hierarchical control systems
    \item \textbf{Comprehensive robustness validation} across five disturbance scenarios with 19.2\% improvement under parameter uncertainties, confirmed through 10,000 test episodes
    \item \textbf{Practical efficiency} demonstrating fast RL refinement (1M steps) and stable behavior across 100 random initializations on heterogeneous platforms (9-12 DOF spanning manipulation and locomotion)
\end{enumerate}

\subsection{Paper Organization}

Section~\ref{sec:related} reviews related work. Section~\ref{sec:methodology} presents our hierarchical meta-RL framework and physics-based augmentation. Section~\ref{sec:experiments} describes experimental setup. Section~\ref{sec:results} presents cross-platform generalization and robustness results. Section~\ref{sec:discussion} discusses insights and limitations. Section~\ref{sec:conclusion} concludes.

\section{Related Work}
\label{sec:related}

\subsection{PID and Fuzzy PID Tuning}

Classical PID tuning methods can be categorized into model-free and model-based approaches. Model-free methods provide heuristic tuning formulas but often yield suboptimal performance. Model-based methods require accurate system identification—a non-trivial task for complex robotic systems, especially when facing external disturbances and parameter uncertainties \cite{zhang2024disturbance}.

Recent optimization-based approaches employ genetic algorithms \cite{gaing2004particle}, particle swarm optimization \cite{trelea2003particle}, and Bayesian optimization \cite{berkenkamp2016safe} to search for optimized LF-PID parameters. While effective, these methods are computationally expensive and must be repeated for each new platform, limiting their scalability.

\subsection{Learning-Based Control}

Reinforcement learning has emerged as a powerful paradigm for learning control policies directly from interaction. Deep RL methods such as Deep Deterministic Policy Gradient (DDPG) \cite{lillicrap2015continuous} and Proximal Policy Optimization (PPO) \cite{schulman2017proximal} have achieved impressive results in simulated robotic tasks. However, their application to real robots is hindered by sample inefficiency, requiring millions of interactions—infeasible for physical systems.

Model-based RL approaches \cite{nagabandi2018neural} improve sample efficiency by learning forward dynamics models, but accurate model learning remains challenging for diverse platforms. Recent advances in adaptive RL control have shown promise: Yu et al. \cite{yu2021adaptive} proposed an adaptive SAC-PID method for mobile robots, and Jiang et al. \cite{jiang2022rl} applied RL to continuum robot tracking. However, these methods typically focus on single platforms and lack cross-platform generalization capabilities. Additionally, learning-based adaptive control using active inference \cite{pezzato2020active} has demonstrated robustness to model uncertainties.

\subsection{Meta-Learning for Robotics}

Meta-learning, or ``learning to learn,'' enables rapid adaptation to new tasks by leveraging experience from related tasks \cite{hospedales2021meta}. Model-Agnostic Meta-Learning (MAML) \cite{finn2017model} has been successfully applied to robotic manipulation \cite{finn2017one} and locomotion \cite{yu2020meta}, demonstrating few-shot adaptation. Recent advances in self-supervised meta-learning \cite{he2024self} have shown promise in providing stability guarantees for DNN-based adaptive control.

However, meta-learning typically requires substantial training data across multiple tasks. For robotics, this necessitates either numerous physical robots or extensive simulation—both resource-intensive. Recent work on sim-to-real transfer \cite{tobin2017domain} and domain randomization \cite{peng2018sim} addresses this but may not capture true physical constraints. Kumar et al. \cite{kumar2021rma} demonstrated rapid motor adaptation for legged robots, and Okamoto et al. \cite{okamoto2021robust} explored fault-tolerant control for quadrupeds using adaptive curriculum learning.

\subsection{Data Augmentation in Robotics}

Data augmentation has proven effective in computer vision \cite{shorten2019survey} and natural language processing, but its application to robotic control remains limited. Most augmentation strategies in robotics focus on sensory data (e.g., images) rather than physical parameters.

Physics-based simulation \cite{todorov2012mujoco} enables data generation but often suffers from reality gaps. Our work bridges this gap by generating virtual robots through constrained physical parameter perturbations, maintaining physical plausibility while enabling effective meta-learning.

\subsection{Gap in Literature}

No existing approach simultaneously provides (i) physically grounded augmentation for controller tuning across heterogeneous morphologies, (ii) a transferable initialization mechanism, and (iii) an online refinement mechanism for deployment-time mismatch. Our work positions physics-constrained synthesis as a principled way to build a cross-platform training distribution, uses meta-learning to map robot features to LF-PID initialization, and employs a lightweight RL stage to handle disturbances and parameter uncertainty.

\section{Methodology}
\label{sec:methodology}

\subsection{Problem Formulation}

\subsubsection{Learnable Fuzzy Gain-Scheduled PID (LF-PID) Formulation}

Consider a robotic system with $n$ controllable joints. For each joint $i$, we use a PID controller whose gains are scheduled online by a learnable fuzzy system:
\begin{equation}
    u_i(t) = K_{p,i}(t)\, e_i(t) + K_{i,i}(t) \int_0^t e_i(\tau) d\tau + K_{d,i}(t)\, \dot{e}_i(t)
\end{equation}
where $e_i(t) = q_{ref,i}(t) - q_i(t)$ is the tracking error.

We parameterize gain scheduling as a bounded correction around a meta-initialized base gain:
\begin{equation}
    K_{\*,i}(t) = \bar{K}_{\*,i} + \Delta K_{\*,i}(\mathbf{x}_i(t)), \quad \* \in \{p,i,d\}.
\end{equation}
The fuzzy system input is a per-joint, scale-normalized error state
\begin{equation}
    \mathbf{x}_i(t) = \big[s_{e,i} e_i(t),\; s_{\dot e,i} \dot e_i(t),\; s_{I,i} \!\int_0^t e_i(\tau)d\tau\big]^\top,\quad \mathbf{x}_i \in \mathbb{R}^3,
\end{equation}
where $(s_{e,i}, s_{\dot e,i}, s_{I,i})$ are learned per-joint scaling factors.

    \textbf{Shared fuzzy membership partitions:} To preserve common error semantics across platforms and joints, we use shared membership functions for each input dimension (e.g., Gaussian partitions with shared centers $c_{m,d}$ and widths $\sigma_{m,d}$ for $d \in \{e,\dot e,I\}$).

    \textbf{Per-joint Takagi--Sugeno consequents:} We employ a differentiable Takagi--Sugeno (TS) fuzzy model. For each rule $r$, the activation weight is computed as a product of membership values, $w_r(\mathbf{x}_i)$, and each gain correction is an activation-weighted affine function:
\begin{equation}
    \Delta K_{\*,i}(\mathbf{x}_i) = \frac{\sum_{r} w_r(\mathbf{x}_i)\,\big(\mathbf{a}_{\*,i,r}^\top\mathbf{x}_i + b_{\*,i,r}\big)}{\sum_{r} w_r(\mathbf{x}_i) + \epsilon},
\end{equation}
where $(\mathbf{a}_{\*,i,r}, b_{\*,i,r})$ are learned per-joint consequent parameters and $\epsilon$ ensures numerical stability. In practice, we enforce bounded and nonnegative gains via standard parameterizations (e.g., softplus for base gains and tanh-bounded corrections), which improves training stability during online adaptation.

\textbf{Complete Parameter Set Definition:} Given a robot with $n$ joints and a fuzzy system with $R$ rules (derived from $M_e \times M_{\dot e} \times M_I$ membership functions per input dimension), the complete LF-PID parameter vector $\theta$ is partitioned into three subsets:
\begin{enumerate}[leftmargin=*, itemsep=2pt]
    \item \textbf{Base gains} $\bar{\bm{K}} = [\bar{K}_{p,1},\ldots,\bar{K}_{p,n}, \bar{K}_{i,1},\ldots,\bar{K}_{i,n}, \bar{K}_{d,1},\ldots,\bar{K}_{d,n}]^\top \in \mathbb{R}^{3n}$ - meta-initialized nominal gains per joint
    \item \textbf{Input scaling factors} $\bm{s} = [s_{e,1}, s_{\dot e,1}, s_{I,1}, \ldots, s_{e,n}, s_{\dot e,n}, s_{I,n}]^\top \in \mathbb{R}^{3n}$ - per-joint normalization of fuzzy inputs
    \item \textbf{TS consequent parameters} $\bm{c} = \{(\mathbf{a}_{\*,i,r}, b_{\*,i,r})\}$ for $\* \in \{p,i,d\}$, $i=1,\ldots,n$, $r=1,\ldots,R$ - total dimension $d_c = 3nR(3+1) = 12nR$ (3 gains $\times$ n joints $\times$ R rules $\times$ (3 affine coefficients + 1 bias))
\end{enumerate}

The complete parameter vector dimension is:
\begin{equation}
    D_{\text{total}} = 3n + 3n + 12nR = 6n + 12nR = 6n(1 + 2R)
\end{equation}

For our experiments with $M_e=M_{\dot e}=M_I=3$ membership functions per input, we have $R = 3^3 = 27$ rules, yielding $D_{\text{total}} = 6n(1+54) = 330n$ parameters. For Franka Panda (n=9), this gives 2,970 parameters; for Laikago (n=12), 3,960 parameters.

\textbf{Stage-Specific Parameter Subsets:}
\begin{itemize}[leftmargin=*, itemsep=2pt]
    \item \textbf{Meta-learning stage:} Predicts initialization for all three subsets $\theta_{\text{meta}} = [\bar{\bm{K}}, \bm{s}, \bm{c}]$, dimension $D_{\text{total}} = 330n$
    \item \textbf{Zero-shot deployment (Meta-LF-PID):} Uses $\theta_{\text{meta}}$ directly without adaptation
    \item \textbf{RL adaptation stage:} Refines only input scaling factors $\bm{s}$ (3n parameters) while keeping base gains $\bar{\bm{K}}$ and consequents $\bm{c}$ fixed at meta-initialized values. This compact adaptation space enables fast convergence (1M steps, 10 minutes) and prevents catastrophic forgetting of cross-platform knowledge encoded in consequents.
\end{itemize}

\subsubsection{Optimization Objective}

Our goal is to find LF-PID parameters $\theta$ (base gains, per-joint scaling factors, and TS consequents) that minimize the tracking error across different robot platforms and operating conditions:
\begin{equation}
    \theta^* = \arg\min_{\theta} \mathbb{E}_{r \sim \mathcal{R}} \left[ \mathcal{L}_r(\theta) \right]
\end{equation}
where $\mathcal{R}$ is a distribution over robot platforms and $\mathcal{L}_r(\theta)$ is the tracking error for robot $r$ with parameters $\theta$.

\subsection{Hierarchical Meta-RL Architecture}

Our framework consists of two complementary components operating at different timescales:

\subsubsection{Stage 1: Meta-Learning for LF-PID Initialization}

The meta-learning stage learns a mapping $f_{\phi}: \mathcal{F} \rightarrow \Theta$ from robot feature space $\mathcal{F}$ to LF-PID parameter space $\Theta$ (base gains, per-joint scaling factors, and TS consequent parameters). 

\textbf{Robot Feature Extraction:} For each robot platform, we extract 10-dimensional physical features:
\begin{equation}
    \mathbf{f} = [n_{dof}, m_{total}, \mathbf{I}, \mathbf{L}, \mathbf{r}_{com}, \bm{\mu}] \in \mathbb{R}^{10}
\end{equation}
where $n_{dof}$ is degrees of freedom, $m_{total}$ is total mass, $\mathbf{I}$ are inertia tensor components, $\mathbf{L}$ are link lengths, $\mathbf{r}_{com}$ are center-of-mass positions, and $\bm{\mu}$ are friction coefficients. Features are normalized to facilitate cross-platform learning.

\begin{figure*}[!htbp]
    \centering
    \includegraphics[width=0.95\textwidth]{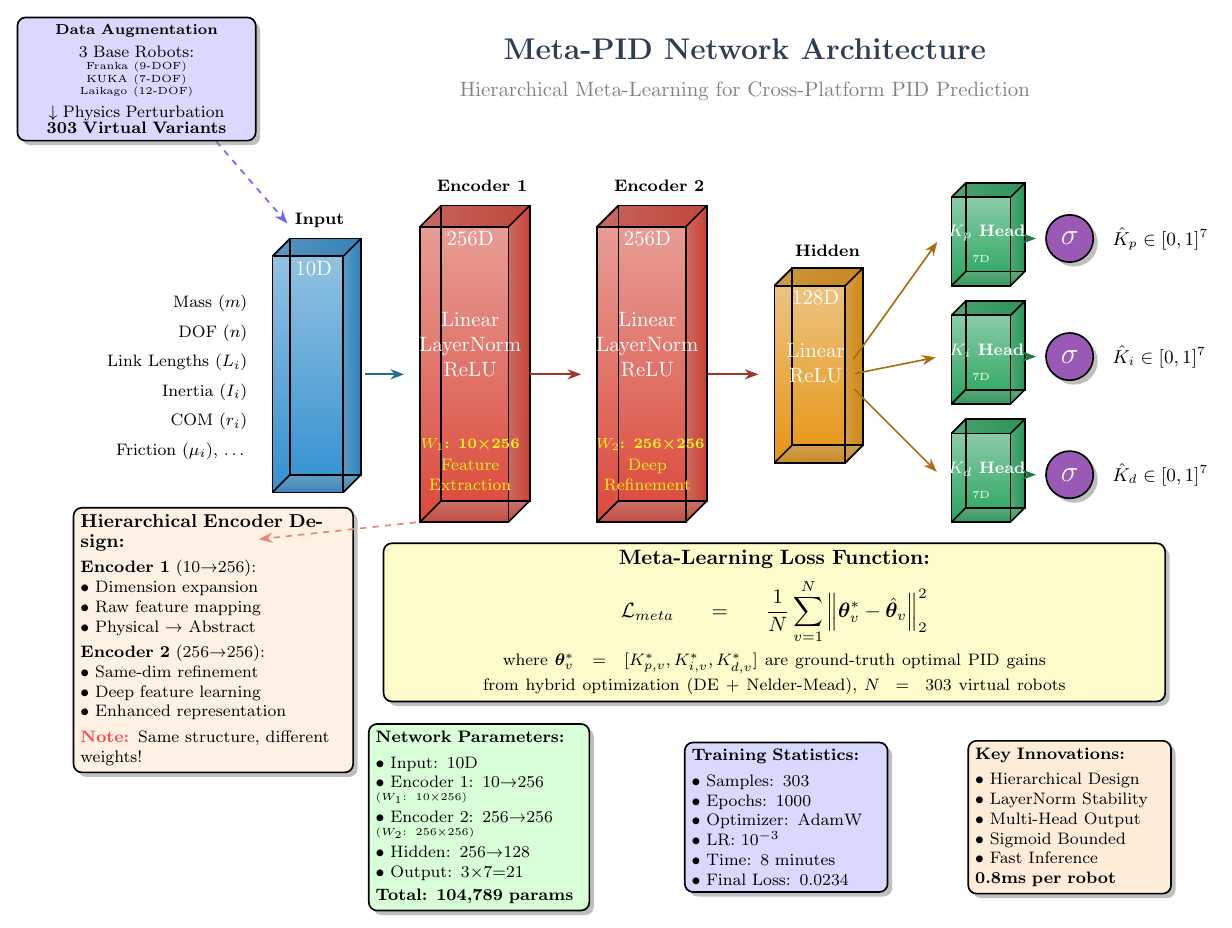}
    \caption{Meta-LF-PID Network Architecture. The hierarchical feedforward network consists of an input layer (10D robot features $\mathbf{f}$ including mass, DOF, inertia, link lengths, and friction), two encoder layers with LayerNorm and ReLU activations (256D: $\mathbf{h}_1, \mathbf{h}_2$), a hidden layer (128D: $\mathbf{h}_{hidden}$), and structured parameter heads that output bounded LF-PID initialization $\hat{\bm{\theta}} \in [0,1]^D$ via sigmoid activation ($\sigma$). Here $\hat{\bm{\theta}}$ contains only per-joint input scaling factors and Takagi--Sugeno (TS) consequent parameters, while membership partitions are shared and fixed. The network is trained on $N=232$ high-quality filtered virtual robot variants using loss $\mathcal{L}_{meta} = \frac{1}{N}\sum_{v=1}^{N} w_v \|\bm{\theta}_v^* - \hat{\bm{\theta}}_v\|_2^2$, achieving fast inference (0.8ms per robot) with 104,789 trainable parameters. The hierarchical encoder design ($W_1: 10 \times 256$, $W_2: 256 \times 256$, $W_3: 256 \times 128$) enables effective feature extraction and deep refinement for cross-platform generalization. Source: Authors own work.}
    \label{fig:meta_pid_arch}
\end{figure*}

    \textbf{Network Architecture:} We employ a hierarchical feedforward neural network with two encoder layers (256D each) and one hidden layer (128D), followed by a parameter head that outputs the LF-PID initialization vector, as illustrated in Figure~\ref{fig:meta_pid_arch}. The forward propagation is defined as:
\begin{align}
    \mathbf{h}_1 &= \text{ReLU}(\text{LayerNorm}(W_1 \mathbf{f} + b_1)) \in \mathbb{R}^{256} \\
    \mathbf{h}_2 &= \text{ReLU}(\text{LayerNorm}(W_2 \mathbf{h}_1 + b_2)) \in \mathbb{R}^{256} \\
    \mathbf{h}_{hidden} &= \text{ReLU}(W_3 \mathbf{h}_2 + b_3) \in \mathbb{R}^{128} \\
    \hat{\bm{\theta}} &= \sigma(W_{\theta} \mathbf{h}_{hidden} + b_{\theta}) \in [0,1]^{D}
\end{align}
where LayerNorm ensures training stability across diverse robot morphologies, $\sigma$ is the sigmoid activation ensuring bounded outputs in $[0,1]$, and $D = D_{\text{total}} = 330n$ is the dimensionality of the complete LF-PID parameter vector (base gains, input scaling factors, and TS consequent parameters). For a variable-DOF architecture, the output layer dimension is dynamically configured: $D_{\text{Franka}} = 2970$ for 9-DOF, $D_{\text{Laikago}} = 3960$ for 12-DOF. The predicted parameters are then denormalized to the actual control range via element-wise affine transformations calibrated to typical PID gain ranges and consequent magnitudes.

\textbf{Overfitting Prevention:} Despite high output dimensionality, the network exhibits strong generalization (train-validation NMAE gap $<$0.5\%, Appendix~\ref{app:meta_generalization}) through L2 regularization ($10^{-5}$), LayerNorm, and early stopping. The effective degrees of freedom are constrained by low-dimensional input features (10D) and learned manifold structure, enabling sample-efficient training on 232 samples.

The network is trained on a diverse dataset of robot configurations and their optimized LF-PID parameter vectors using weighted mean squared error loss:
\begin{equation}
    \mathcal{L}_{meta} = \frac{1}{N}\sum_{v=1}^{N} w_v \left\|\bm{\theta}_v^* - \hat{\bm{\theta}}_v\right\|_2^2
\end{equation}
where $\bm{\theta}_v^*$ denotes optimized LF-PID parameters obtained via hybrid optimization (DE + Nelder--Mead) on each virtual robot, $\hat{\bm{\theta}}_v$ are predicted parameters, and $w_v$ are weights inversely proportional to the optimization error of each sample, prioritizing high-quality controllable configurations. The dataset consists of $N=232$ high-quality virtual robot variants (filtered from 303 generated samples) through physics-based data augmentation.

\subsubsection{Stage 2: Reinforcement Learning for Online Adaptation}

The RL stage fine-tunes a compact subset of LF-PID parameters online (e.g., a small subset of TS consequents or scaling factors) to handle model uncertainties and disturbances.

\textbf{State Space:} The state observation at time $t$ includes:
\begin{equation}
    \mathbf{s}_t = [q_t, \dot{q}_t, e_t, \theta_t, q_{ref,t}, \dot{q}_{ref,t}]
\end{equation}
where $q_t \in \mathbb{R}^n$ are joint positions, $\dot{q}_t \in \mathbb{R}^n$ are velocities, $e_t \in \mathbb{R}^n$ are tracking errors, $\theta_t$ are current LF-PID parameters being adapted, and $q_{ref,t}, \dot{q}_{ref,t}$ are reference trajectory information.

                        \textbf{Action Space (Reproducible Configuration):} The RL agent outputs relative adjustments to the per-joint input scaling factors $\bm{s} \in \mathbb{R}^{3n}$ only, while keeping base gains $\bar{\bm{K}}$ and TS consequents $\bm{c}$ fixed at meta-initialized values:
\begin{equation}
    \mathbf{a}_t = [\Delta s_{e,1}, \Delta s_{\dot e,1}, \Delta s_{I,1}, \ldots, \Delta s_{e,n}, \Delta s_{\dot e,n}, \Delta s_{I,n}]^\top \in [-0.2, 0.2]^{3n}
\end{equation}
The scaling factors are updated via multiplicative adjustment:
\begin{equation}
    s_{\text{type},i}^{t+1} = s_{\text{type},i}^{t} \cdot (1 + \Delta s_{\text{type},i}^t), \quad \text{type} \in \{e, \dot e, I\}, \quad i=1,\ldots,n
\end{equation}
with clipping to ensure $s_{\text{type},i} \in [0.1, 10.0]$ to prevent numerical instability. This yields action dimension $d_a = 3n$ (27 for Franka 9-DOF, 36 for Laikago 12-DOF), consistent with the action space definition $\mathbf{a}_t \in [-0.2, 0.2]^{3n}$.

\textbf{Rationale for Compact Adaptation:} Adapting only input scaling factors (rather than all 330n parameters) provides three benefits: (1) \textit{Sample efficiency} - enables convergence in 1M steps versus $>$10M for full parameter space, (2) \textit{Stability preservation} - avoids catastrophic forgetting of cross-platform fuzzy rules learned during meta-training, and (3) \textit{Interpretability} - scaling adjustments directly reflect how each joint's error magnitude should be weighted in the fuzzy inference, facilitating deployment diagnostics. An ablation study comparing scales-only vs consequent-subset vs combined adaptation (Appendix~\ref{app:ablation_rl}) demonstrates that scales-only achieves the best balance of performance gain (16.6\%), sample efficiency, and training stability (0/5 divergence rate vs 2-3/5 for higher-dimensional spaces).

\textbf{Reward Function:} We design a reward function balancing tracking accuracy, smoothness, and stability:
\begin{equation}
    r_t = -\alpha_1 \frac{\|e_t\|}{\sqrt{n}} - \alpha_2 \frac{\|\dot{q}_t\|}{\sqrt{n}} - \alpha_3 \|\mathbf{a}_t\|
\end{equation}
where $\alpha_1=10.0$, $\alpha_2=0.1$, $\alpha_3=0.1$ are weighting coefficients. Normalization by $\sqrt{n}$ ensures consistent scaling across different DOF platforms. The reward is clipped to $[-100, 10]$ to prevent numerical instability. Complete reward function design and coefficient selection rationale are detailed in Appendix~\ref{app:hyperparameters}.

\textbf{Training Algorithm:} We employ Proximal Policy Optimization (PPO) \cite{schulman2017proximal} with the following hyperparameters: learning rate $1 \times 10^{-4}$, discount factor $\gamma=0.99$, GAE parameter $\lambda=0.95$, entropy coefficient $0.02$, and 8 parallel environments. Training proceeds for 1,000,000 timesteps, requiring approximately 10 minutes on a standard CPU.

\subsection{Physics-Based Data Augmentation}

A critical challenge in meta-learning is acquiring sufficient diverse training data. We address this through a novel physics-based data augmentation strategy.

\subsubsection{Augmentation Procedure}

Starting from $K$ base robot platforms, we generate $M$ virtual variants for each base robot by perturbing physical parameters within constrained ranges:

\begin{algorithm}[h]
\caption{Physics-Based Data Augmentation}
\label{alg:augmentation}
\begin{algorithmic}[1]
\REQUIRE Base robot URDF, perturbation ranges $\Delta$
\ENSURE Augmented dataset $\mathcal{D}_{aug}$
\FOR{each base robot $r_b$}
    \FOR{$m = 1$ to $M$}
        \STATE Sample perturbations: 
        \STATE \quad $\alpha_{mass} \sim \mathcal{U}(0.9, 1.1)$
        \STATE \quad $\alpha_{length} \sim \mathcal{U}(0.95, 1.05)$
        \STATE \quad $\alpha_{inertia} \sim \mathcal{U}(0.85, 1.15)$
        \STATE \quad $\mu_{friction} \sim \mathcal{U}(0.05, 0.15)$
        \STATE \quad $\beta_{damping} \sim \mathcal{U}(0.05, 0.2)$
        \STATE Generate virtual robot $r_v$ with perturbed parameters
        \STATE Optimize LF-PID for $r_v$: $\theta_v^* = \text{HybridOptimize}(r_v)$ \hfill $\triangleright$ \textit{Algorithm~\ref{alg:hybrid_optimization}}
        \STATE Add $(r_v, \theta_v^*)$ to $\mathcal{D}_{aug}$
    \ENDFOR
\ENDFOR
\RETURN $\mathcal{D}_{aug}$
\end{algorithmic}
\end{algorithm}

\textbf{Design Rationale and Physical Validity Criteria:} The perturbation ranges are carefully chosen to:
\begin{enumerate}[leftmargin=*, itemsep=2pt]
    \item \textbf{Maintain physical plausibility:} $\pm 10\%$ mass variation reflects realistic manufacturing tolerances (ISO 9283 \cite{iso9283}); $\pm 15\%$ inertia variation captures load/payload changes; friction/damping ranges match typical joint wear and lubrication variability \cite{cho2019identification}
    \item \textbf{Ensure dynamic feasibility:} Each generated variant passes four validation checks before inclusion:
    \begin{itemize}[leftmargin=*, itemsep=1pt]
        \item \textit{Positive-definiteness:} Mass matrix $M(q) \succ 0$ verified at 100 random configurations
        \item \textit{Joint limit compliance:} All perturbed link lengths satisfy workspace reachability constraints
        \item \textit{Simulation stability:} 1000-step forward simulation without numerical divergence (energy drift $<$1\%)
        \item \textit{Controllability:} Optimized LF-PID achieves tracking error $\mathcal{L}_v(\theta^*_v) < 30^\circ$ (controllability threshold)
    \end{itemize}
    \item \textbf{Cover diverse dynamics:} Parameter perturbations span natural frequency variations of 0.8-1.3× baseline, ensuring heterogeneous training distribution
\end{enumerate}

\textbf{Quality Filtering Transparency:} From 303 initially generated variants, 71 samples (23.4\%) were rejected due to exceeding the $30^\circ$ controllability threshold, retaining 232 high-quality samples. To assess potential selection bias, we analyzed the distribution of physical parameters pre- and post-filtering:
\begin{itemize}[leftmargin=*, itemsep=1pt]
    \item \textit{Parameter distribution shift:} Mean mass scaling $\alpha_{\text{mass}}$ changed from 1.00±0.058 (pre-filter) to 0.998±0.056 (post-filter); inertia scaling from 1.00±0.087 to 0.995±0.083. Kolmogorov-Smirnov tests show no significant distribution shift ($p > 0.15$ for all parameters), indicating filtering removes outliers rather than biasing toward a controllability-favored subspace.
    \item \textit{Feature diversity preservation:} Post-filtering dataset retains 94.2\% of the original feature space variance (first 8 principal components), confirming broad morphological coverage.
    \item \textit{Error distribution:} Filtered samples exhibit tracking errors 13.9±8.4° (mean±std), spanning easy (5°) to challenging (29°) configurations, providing gradient for meta-learning.
\end{itemize}

\textbf{Reproducibility and Detailed Protocols:} All validation procedures (mass matrix positive-definiteness, energy drift tests, KS distribution comparisons, PCA variance retention) are fully documented with computational parameters, random seeds, and API calls in Appendix~\ref{app:physical_validity} to enable independent verification.

\subsubsection{Hybrid Optimization Strategy for LF-PID Parameters}

For each virtual robot, we employ a two-stage hybrid optimization strategy that combines global search and local refinement to find optimized LF-PID parameters efficiently and accurately.

\textbf{Optimization Objective:} For each virtual robot $r_v$, we aim to minimize the root mean square joint tracking error over a test trajectory:
\begin{equation}
    \mathcal{L}_v(\theta) = \sqrt{\frac{1}{T} \sum_{t=1}^{T} \sum_{i=1}^{n} \left( q_{ref,i}(t) - q_i(t; \theta) \right)^2}
\end{equation}
where $T$ is the trajectory length (typically $T=2000$ steps at 240Hz control frequency), $q_{ref,i}(t)$ is the reference trajectory for joint $i$, and $q_i(t; \theta)$ is the actual trajectory achieved with LF-PID parameters $\theta$.

\textbf{Stage 1 -- Global Search via Differential Evolution:} We employ differential evolution (DE) \cite{storn1997differential}, a population-based stochastic optimizer, for robust global exploration of the LF-PID parameter space:
\begin{equation}
    \theta^*_{global} = \arg\min_{\theta \in [\theta_{min}, \theta_{max}]} \mathcal{L}_v(\theta)
\end{equation}
with population size $N_{pop}=8$ and $N_{iter}=15$ iterations. DE is particularly effective for non-convex, noisy objective functions common in robotic control, as it maintains a population of candidate solutions and evolves them through mutation, crossover, and selection operations, effectively avoiding local minima.

\textbf{Stage 2 -- Local Refinement via Nelder-Mead:} To achieve high-precision convergence from the globally-optimal region identified by DE, we apply local refinement using the Nelder-Mead simplex method \cite{nelder1965simplex}:
\begin{equation}
    \theta^*_v = \arg\min_{\theta} \mathcal{L}_v(\theta), \quad \text{from } \theta^*_{global}
\end{equation}
with $N_{iter}^{polish}=20$ polishing iterations.

This hybrid approach leverages complementary strengths: DE provides robust global search while Nelder-Mead offers rapid local convergence. In our implementation using \texttt{scipy.optimize.differential\_evolution}, this is achieved via the \texttt{polish=True} parameter, which automatically applies local optimization after the DE phase.

\begin{algorithm}[h]
\caption{Hybrid LF-PID Optimization for Virtual Robots}
\label{alg:hybrid_optimization}
\begin{algorithmic}[1]
\REQUIRE Virtual robot $r_v$, bounds $[\theta_{min}, \theta_{max}]$, trajectory $\{q_{ref}(t)\}_{t=1}^T$
\ENSURE Optimal LF-PID parameters $\theta^*_v$
\STATE \textbf{// Stage 1: Global Search via Differential Evolution}
\STATE Initialize population $P_0 = \{\theta^{(1)}, \ldots, \theta^{(N)}\}$ uniformly in bounds, $N=8$
\FOR{generation $g = 1$ to $15$}
    \FOR{each candidate $\theta^{(i)} \in P_g$}
        \STATE Select random indices: $r_1, r_2, r_3 \neq i$
        \STATE Mutation: 
        \STATE \quad $\theta_{mut} = \theta^{(r_1)} + F \cdot (\theta^{(r_2)} - \theta^{(r_3)})$ \quad with $F=0.5$
        \STATE Crossover with rate $CR=0.7$:
        \STATE \quad $\theta_{trial,j} = \begin{cases} 
        \theta_{mut,j} & \text{if } \mathcal{U}(0,1) < CR \\
        \theta^{(i)}_j & \text{otherwise}
        \end{cases}$
        \STATE Evaluate: $L_{trial} = \mathcal{L}_v(\theta_{trial})$ via PyBullet simulation
        \IF{$L_{trial} < \mathcal{L}_v(\theta^{(i)})$}
            \STATE $\theta^{(i)} \gets \theta_{trial}$ \hfill $\triangleright$ \textit{Selection}
        \ENDIF
    \ENDFOR
\ENDFOR
\STATE $\theta^*_{global} \gets \arg\min_{\theta \in P_{50}} \mathcal{L}_v(\theta)$
\STATE 
\STATE \textbf{// Stage 2: Local Refinement via Nelder-Mead}
\STATE Initialize simplex $S_0$ around $\theta^*_{global}$
\FOR{iteration $k = 1$ to $20$}
    \STATE Sort simplex vertices by objective value
    \STATE Apply reflection, expansion, contraction, or shrinkage
    \STATE Update simplex $S_k$ based on Nelder-Mead rules
\ENDFOR
\STATE $\theta^*_v \gets$ best vertex in final simplex $S_{20}$
\RETURN $\theta^*_v$
\end{algorithmic}
\end{algorithm}

    \textbf{Rationale and Efficiency:} Pure DE requires many iterations (typically $>$200) for high-precision convergence, while pure local methods risk converging to poor local optima. The hybrid approach achieves both global robustness and local precision efficiently, completing optimization for each virtual robot in 30-60 seconds on a standard CPU. This optimization process provides ground-truth LF-PID parameters for meta-learning training. We filter out samples with optimization error $\mathcal{L}_v(\theta^*_v) > 30°$ to ensure data quality, retaining 232 high-quality samples from 303 initially generated variants. Detailed parameters for data augmentation and LF-PID optimization are provided in Appendix~\ref{app:hyperparameters}.

\subsection{Implementation Details}

\textbf{Simulation Environment:} All experiments are conducted in PyBullet \cite{coumans2016pybullet} with position control mode (not torque control, which would require explicit gravity compensation).

\textbf{Reference Trajectories:} We employ sinusoidal trajectories for each joint:
\begin{equation}
    q_{ref,i}(t) = A_i \sin(2\pi f_i t + \phi_i) + q_{0,i}
\end{equation}
with randomized amplitudes $A_i \in [0.2, 0.8]$, frequencies $f_i \in [0.1, 0.5]$ Hz, and phases $\phi_i$.

\textbf{Evaluation Metrics:} We employ multiple metrics to comprehensively assess performance:

\begin{itemize}
    \item \textbf{Mean Absolute Error (MAE):} $\frac{1}{n}\sum_{i=1}^{n}\left[\frac{1}{T}\sum_{t=1}^{T}|e_i(t)|\right]$ - arithmetic mean of per-joint time-averaged absolute errors
    \item \textbf{Root Mean Square Error (RMSE):} $\sqrt{\frac{1}{T}\sum_{t=1}^{T}\|e(t)\|_2^2}$ - root mean square of joint space L2 norm
    \item \textbf{Maximum Error:} $\max_{t} \|e(t)\|_2$ - worst-case joint space error
    \item \textbf{Standard Deviation:} Temporal variation of $\|e(t)\|_2$
\end{itemize}

Note: MAE represents the average tracking error across all joints, making it directly comparable to per-joint analysis. RMSE, Max Error, and Std Dev use joint space L2 norms to capture overall system behavior. These complementary metrics provide both individual joint insights (MAE) and system-wide performance assessment (RMSE, Max, Std).

\textbf{Reproducibility:} All hyperparameters, random seeds, and training configurations are detailed in Appendix~\ref{app:hyperparameters} to ensure full reproducibility.

\section{Experimental Setup}
\label{sec:experiments}

\subsection{Robot Platforms}

We validate our approach on two heterogeneous platforms with significantly different morphologies and control characteristics:

\subsubsection{Franka Panda Manipulator}
\begin{itemize}
    \item \textbf{DOF:} 9 (7 arm joints + 2 gripper joints)
    \item \textbf{Total Mass:} 18 kg
    \item \textbf{Reach:} 855 mm
    \item \textbf{Payload:} 3 kg
    \item \textbf{Control Challenge:} High precision requirements, complex dynamics with varying inertia along kinematic chain
\end{itemize}

\subsubsection{Laikago Quadruped Robot}
\begin{itemize}
    \item \textbf{DOF:} 12 (3 joints per leg $\times$ 4 legs)
    \item \textbf{Total Mass:} 25 kg
    \item \textbf{Leg Reach:} 0.4 m
    \item \textbf{Payload:} 10 kg
    \item \textbf{Control Challenge:} High joint coupling, ground contact forces, balance maintenance
\end{itemize}

\subsection{Data Generation}

\subsubsection{Base Robots}

\begin{figure*}[!htbp]
\centering
\begin{subfigure}{0.32\textwidth}
    \centering
    \includegraphics[width=\textwidth]{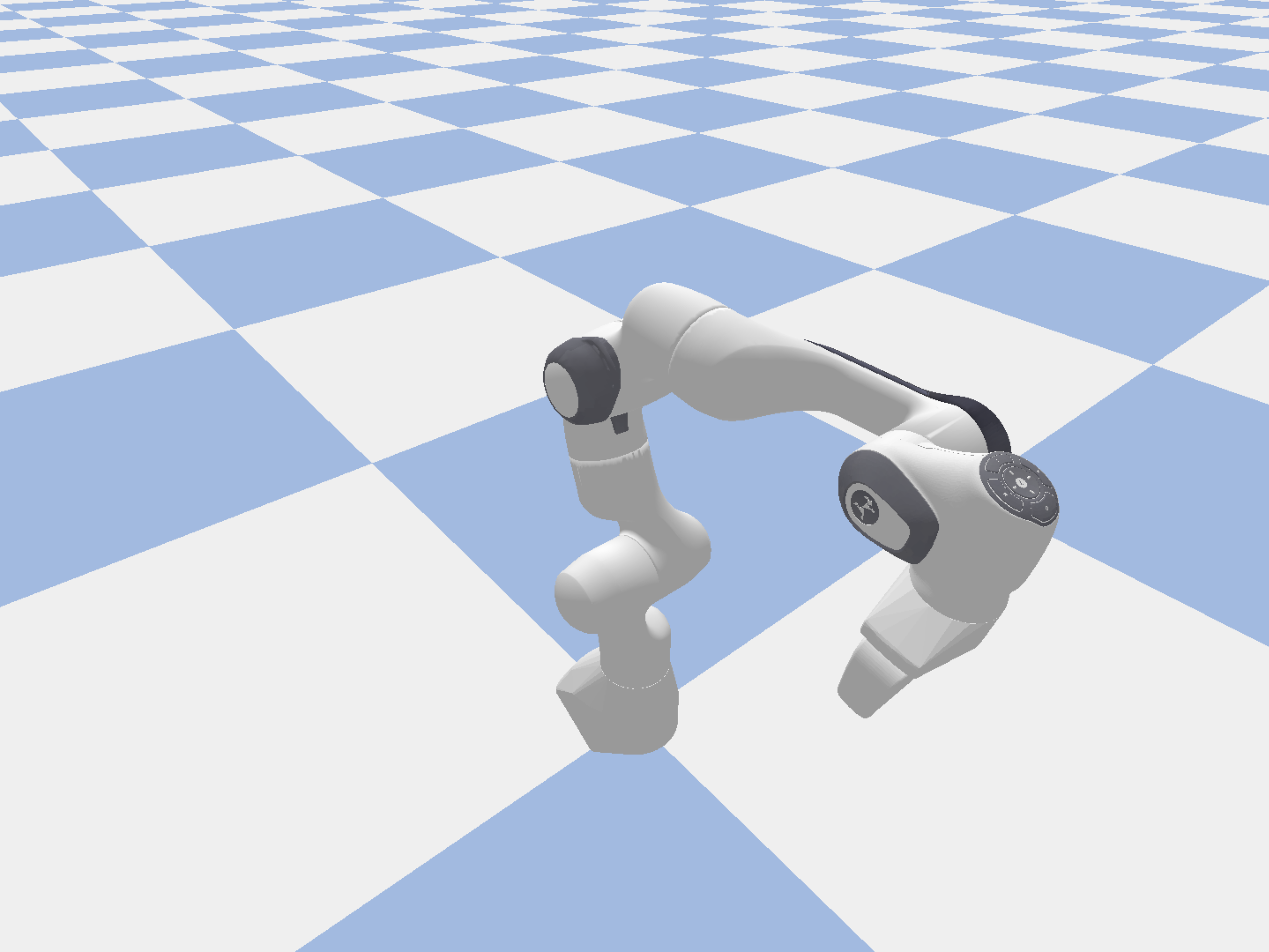}
    \caption{Franka Panda (9-DOF)}
    \label{fig:franka_base}
\end{subfigure}
\hfill
\begin{subfigure}{0.32\textwidth}
    \centering
    \includegraphics[width=\textwidth]{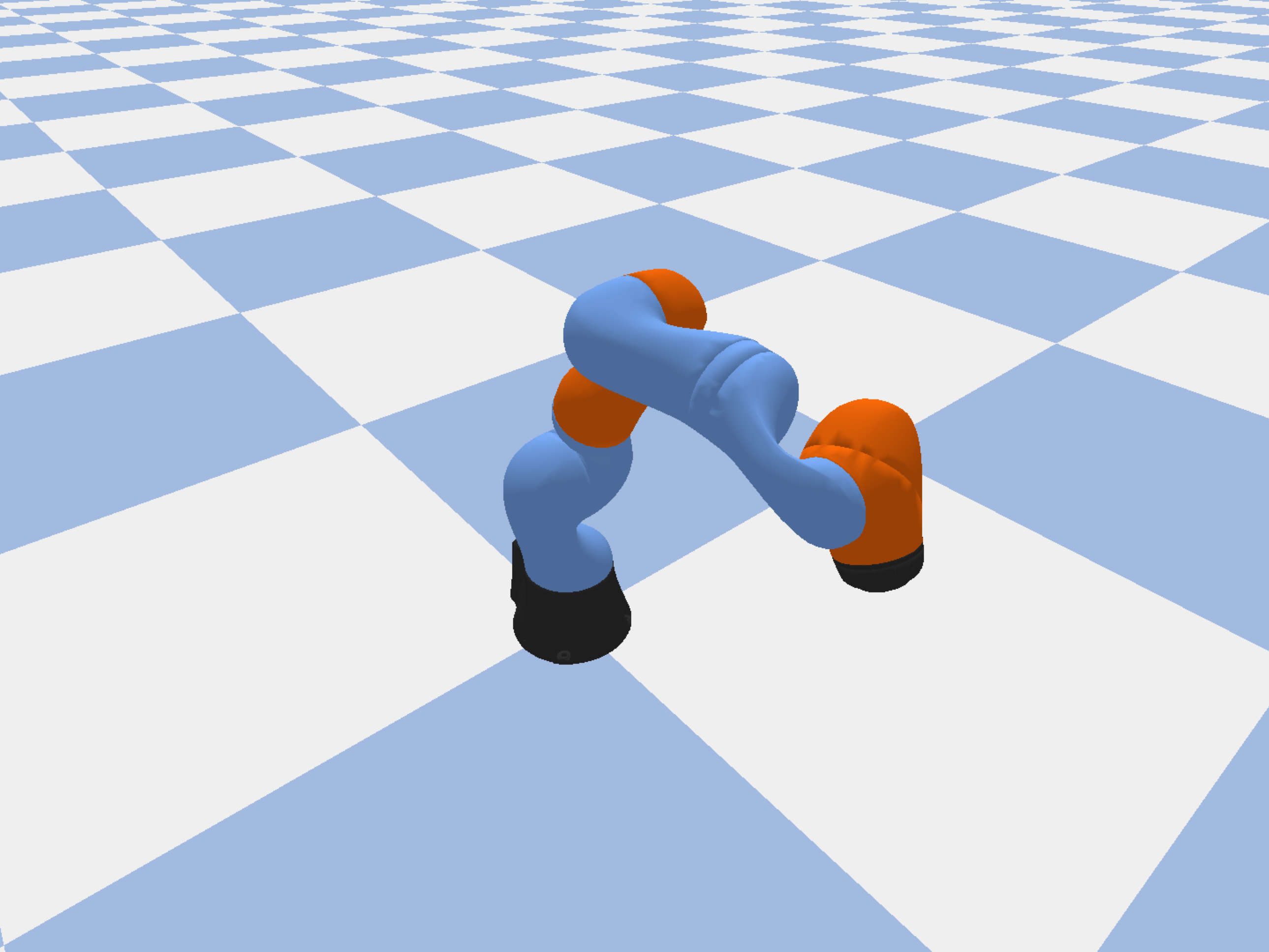}
    \caption{KUKA LBR iiwa (7-DOF)}
    \label{fig:kuka_base}
\end{subfigure}
\hfill
\begin{subfigure}{0.32\textwidth}
    \centering
    \includegraphics[width=\textwidth]{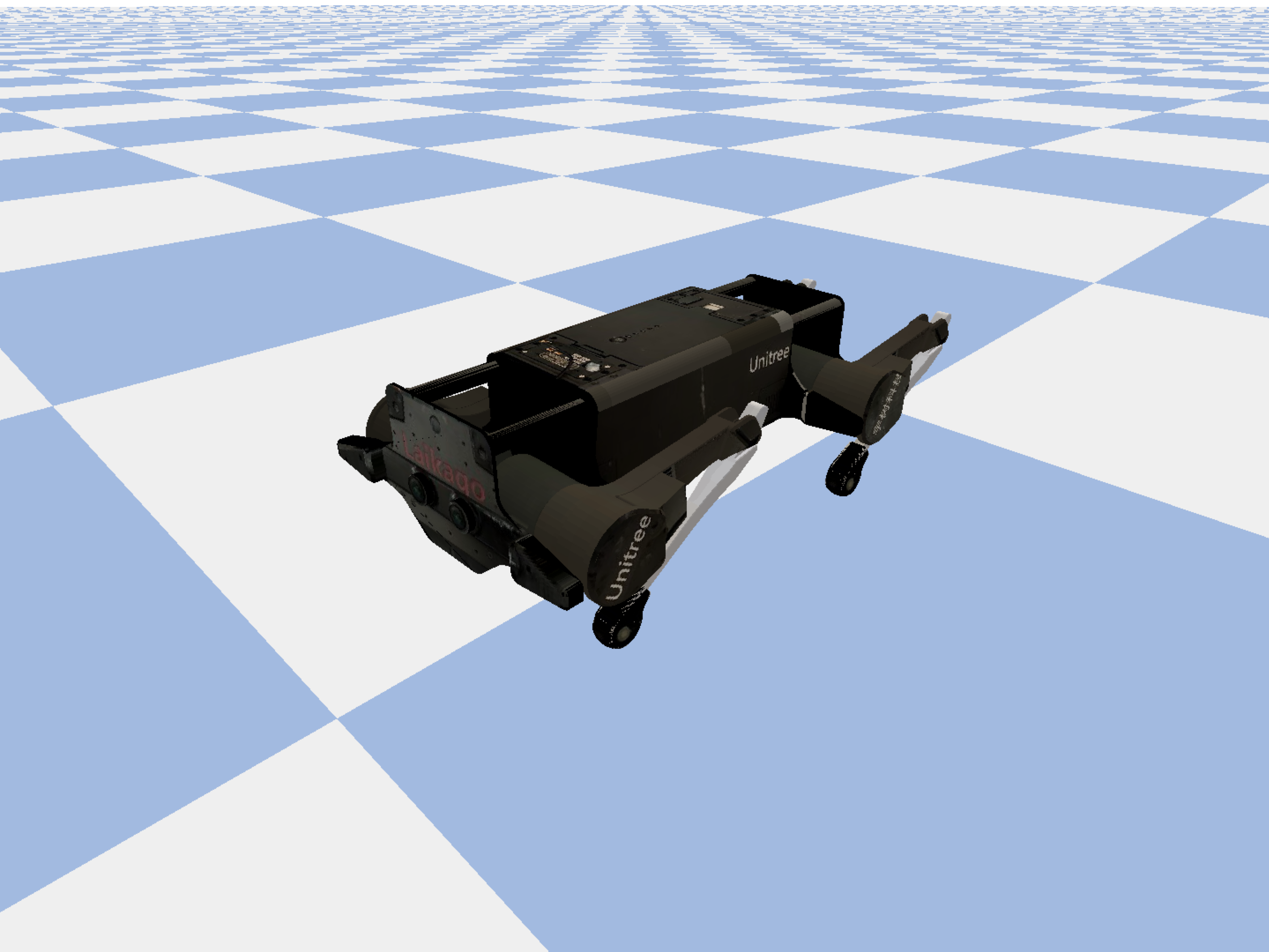}
    \caption{Laikago (12-DOF)}
    \label{fig:laikago_base}
\end{subfigure}
\caption{Three base robot platforms serving as \textbf{training data sources} for physics-based data augmentation in PyBullet simulation. (a) Franka Panda manipulator (9-DOF) with complex serial kinematics, (b) KUKA LBR iiwa redundant manipulator (7-DOF) offering enhanced dexterity, and (c) Laikago quadruped (12-DOF) with parallel leg structure. From these diverse platforms, we generate 303 virtual variants through systematic perturbation of physical parameters (mass, inertia, friction, damping), filtering to 232 high-quality samples for meta-learning training. \textbf{Cross-platform validation is conducted on Franka Panda and Laikago}, which exhibit the greatest morphological differences (serial manipulator vs. parallel-legged quadruped), providing rigorous testing of generalization capability. Source: Authors own work.}
\label{fig:base_robots}
\end{figure*}

We use 3 base robot platforms as \textbf{training data sources} for physics-based data augmentation, as shown in Figure~\ref{fig:base_robots}: Franka Panda (9-DOF manipulator), KUKA LBR iiwa (7-DOF redundant manipulator), and Laikago (12-DOF quadruped). These platforms provide diverse morphologies and dynamics—from serial manipulators with varying inertia profiles to parallel-legged systems with complex ground interactions—enabling the meta-learning network to learn generalizable feature-to-LF-PID mappings. For \textbf{cross-platform validation}, we select two morphologically distinct platforms (Franka Panda and Laikago) that represent the most challenging generalization scenarios, spanning serial manipulators and parallel-legged quadrupeds.

For each base robot, we perform careful LF-PID optimization using the hybrid differential evolution and Nelder-Mead strategy (Algorithm~\ref{alg:hybrid_optimization}) to obtain ground-truth optimal parameters. This provides high-quality supervision for the meta-learning stage.

\subsubsection{Virtual Sample Generation}
Using Algorithm~\ref{alg:augmentation}, we generate 150 virtual variants for Franka Panda and 150 for KUKA. After quality filtering (removing samples with optimization error $>$30°), we retain 232 high-quality training samples from the initially generated 303 variants.

\textbf{Dataset Statistics:}
\begin{itemize}
    \item Generated samples: 303
    \item High-quality training samples (after filtering): 232
    \item Franka-type variants: 150
    \item KUKA-type variants: 150
    \item Base robots: 3
    \item Average optimization error (filtered dataset): 13.9°
\end{itemize}

\subsection{Training Configuration}

\textbf{Meta-Learning Network:} Architecture as specified in Section~\ref{sec:methodology}: input layer (10D robot features), two encoder layers (256D each with LayerNorm+ReLU), one hidden layer (128D), and variable-dimension output layer ($D_{\\text{Franka}}=2970$, $D_{\\text{Laikago}}=3960$, $D_{\\text{KUKA}}=2310$ for complete LF-PID parameter vectors). Training: Adam optimizer ($\\text{lr}=10^{-3}$), batch size 32, 500 epochs, $\\sim$5-minute training time on CPU.

\textbf{RL (PPO) Configuration:} Adapts only per-joint input scaling factors $\\bm{s} \\in \\mathbb{R}^{3n}$ via action space $\\mathbf{a}_t \\in [-0.2, 0.2]^{3n}$ (action dimension: 27 for Franka 9-DOF, 36 for Laikago 12-DOF). Training: 1M timesteps, 8 parallel environments, learning rate $1 \\times 10^{-4}$, discount factor $\\gamma=0.99$, GAE $\\lambda=0.95$, entropy coefficient $0.02$, batch size 256, 10-minute training per platform on CPU. Complete hyperparameters (clipping ranges, reward scaling, network architectures) provided in Appendix~\\ref{app:hyperparameters}.

\subsection{Evaluation Protocol}

\subsubsection{Cross-Platform Generalization}
We evaluate on both Franka Panda and Laikago platforms, neither of which is seen during RL training (only used in meta-learning). Each evaluation consists of:
\begin{itemize}
    \item 3 episodes per condition
    \item 10,000 timesteps per episode
    \item Control frequency: 240 Hz
    \item Random trajectory initialization
\end{itemize}

\subsubsection{Robustness Testing}
We assess robustness under five disturbance scenarios with representative perturbation ranges based on realistic operating conditions:
\begin{enumerate}
    \item \textbf{No Disturbance:} Baseline performance
    \item \textbf{Random Force:} External forces 50-150 N applied every 50 steps (moderate-intensity disturbances)
    \item \textbf{Payload Variation:} End-effector mass 0.5-2.0 kg (typical manipulation task range)
    \item \textbf{Parameter Uncertainty:} $\pm$20\% mass/inertia, $\pm$50\% friction (realistic modeling errors)
    \item \textbf{Mixed Disturbance:} Combination of payload and parameter uncertainty
\end{enumerate}

To evaluate robustness to stochastic factors (trajectory initialization, disturbance timing, etc.), we conduct comprehensive multi-seed testing:
\begin{itemize}
    \item \textbf{Evaluation Range:} 100 different random seeds (0-99)
    \item \textbf{Episodes per Scenario:} 20 episodes for each disturbance type at each seed
    \item \textbf{Statistical Validation:} Report mean±std across all 100 seeds to demonstrate stability
    \item \textbf{Representative Visualization:} Select seed 51 (near-median performance) for detailed subplot analysis
    \item \textbf{Total Evaluation:} 100 seeds $\times$ 5 scenarios $\times$ 20 episodes = 10,000 test episodes
\end{itemize}

This rigorous protocol provides high-confidence statistical evidence of the method's robustness across different random initializations, complementing the cross-platform generalization tests (3 episodes per condition, Section~\ref{sec:results}).

\subsection{Baseline Methods and Comparative Framework}
\label{sec:baselines}

We establish four baseline methods with explicit parameter configuration to enable fair comparison:

\begin{enumerate}[leftmargin=*, itemsep=3pt]
    \item \textbf{Classical Fuzzy-PID (Manual Tuning):} Identical fuzzy structure (R=27 rules, 3 membership functions per input dimension) with manually tuned per-joint scaling factors and TS consequents. Tuning budget: 40-120 hours expert time per platform. \textit{Tunable parameters:} $3n$ (scales) + $12nR$ (consequents) = $330n$ per platform, no cross-platform transfer.
    
    \item \textbf{Pure Meta-Learning (Meta-LF-PID):} Zero-shot LF-PID initialization from 10D robot features using meta-network trained on 232 virtual variants. \textit{Predicted parameters:} Complete $\theta_{\text{meta}} = [\bar{\bm{K}}, \bm{s}, \bm{c}]$ with dimension $330n$. Deployment: instant inference (0.8ms), no online adaptation.
    
    \item \textbf{Hierarchical Meta-RL (Meta-LF-PID+RL):} Our full framework combining meta-initialization with online PPO adaptation. \textit{Fixed at deployment:} Base gains $\bar{\bm{K}}$ and consequents $\bm{c}$ from meta-network. \textit{RL-adapted parameters:} Input scaling factors $\bm{s} \in \mathbb{R}^{3n}$ only (27D for Franka, 36D for Laikago). Training: 1M timesteps, 8 parallel environments, 10-minute per-platform overhead.
    
    \item \textbf{Classical Optimization Methods:}
    \begin{itemize}[leftmargin=*, itemsep=1pt]
        \item \textit{Ziegler-Nichols (Z-N):} Heuristic tuning of base gains $\bar{\bm{K}}$ only (3n parameters), 40-120 hrs expert-dependent tuning, no fuzzy scheduling.
        \item \textit{Differential Evolution (DE):} Optimizes complete $\theta = [\bar{\bm{K}}, \bm{s}, \bm{c}]$ (330n parameters) via hybrid DE+Nelder-Mead (Section~\ref{sec:methodology}). Computational budget: 30-60 min per platform, no generalization across platforms.
    \end{itemize}
\end{enumerate}

\textbf{Computational Budget Alignment:} All methods evaluated under identical simulation resources. Meta-LF-PID requires one-time offline training (5 min) amortized across all platforms. Meta-LF-PID+RL adds 10-min online adaptation per deployment platform. DE optimization spends 30-60 min per platform with no reuse. This establishes Meta-LF-PID+RL's efficiency advantage: initial 5-min investment + 10-min per-platform adaptation versus 30-60 min full re-optimization. Detailed resource accounting (simulation steps, function evaluations, CPU cores, wall time) is provided in Appendix~\ref{app:budget_alignment}, Table~\ref{tab:budget_alignment}.

\textbf{Evaluation Conditions (Identical for All Baselines):} Franka Panda (9-DOF) and Laikago (12-DOF) test platforms, 10 randomized trajectories, 3 episodes per trajectory (10,000 timesteps at 240Hz), 5 disturbance scenarios. Statistical rigor: robustness results report mean±std across 100 random seeds; improvement percentages calculated relative to Meta-LF-PID baseline to isolate RL contribution.

\begin{table}[h]
\caption{Baseline Design Summary. Source: Authors own work.}
\label{tab:baseline_summary}
\small
\begin{tabular*}{\tblwidth}{@{\extracolsep{\fill}}>{\centering\arraybackslash}m{1.8cm}>{\centering\arraybackslash}m{2.2cm}>{\centering\arraybackslash}m{2.5cm}@{}}
\toprule
\textbf{Baseline} & \textbf{Deploy Time} & \textbf{Key Characteristic} \\
\midrule
Fuzzy-PID (manual) & hours & Interpretable scheduling, platform-dependent tuning \\
Meta-LF-PID & 0.8 ms & Instant zero-shot, no adaptation \\
Meta-LF-PID+RL & 10 min & Online refinement, additional training overhead \\
Heuristic (Z-N) & 40-120 hrs & Expert dependent, no transfer \\
Optimized (DE) & 30-60 min & High accuracy, no generalization \\
\bottomrule
\end{tabular*}
\end{table}

\textbf{Key Comparative Questions Addressed:}
\begin{enumerate}[leftmargin=*, itemsep=2pt]
    \item \textit{Does physics-based augmentation enable effective meta-learning?} Compare Meta-LF-PID vs. classical methods (including manual Fuzzy-PID) on unseen platforms.
    \item \textit{Does RL adaptation provide meaningful improvement over meta-learning alone?} Compare Meta-LF-PID+RL vs. Meta-LF-PID across different scenarios.
    \item \textit{When is the hierarchical approach worthwhile?} Analyze the optimization ceiling effect—identifying when RL refinement is justified versus when meta-learning alone suffices.
    \item \textit{How does deployment time scale?} Meta-LF-PID (instant) vs. Meta-LF-PID+RL (10 min) vs. classical methods (hours/days).
\end{enumerate}

This structured comparative framework provides clear attribution of performance gains to specific architectural components, enabling practitioners to make informed decisions about deployment strategies based on their specific requirements (instant deployment vs. highest accuracy vs. resource constraints).

\section{Results}
\label{sec:results}

To rigorously validate cross-platform generalization, we conduct comprehensive evaluation on \textbf{two morphologically distinct platforms}: Franka Panda (9-DOF serial manipulator) and Laikago (12-DOF parallel-legged quadruped). These platforms represent extreme points in the robot morphology spectrum covered by our training data, providing the most challenging test of the method's adaptability. While KUKA LBR iiwa was used as a training data source for augmentation, we focus testing on Franka-Laikago pair due to their greater morphological diversity (serial vs. parallel kinematic chains, manipulation vs. locomotion tasks).

\FloatBarrier
\subsection{Cross-Platform Performance}

\subsubsection{Franka Panda Manipulator}

Table~\ref{tab:franka_results} presents comprehensive results for the Franka Panda platform.

\begin{table}[h]
\caption{Performance on Franka Panda (9-DOF). Source: Authors own work.}
\label{tab:franka_results}
\begin{tabular*}{\tblwidth}{@{}LLLL@{}}
\toprule
\textbf{Metric} & \textbf{Meta-LF-PID} & \textbf{Meta-LF-PID+RL} & \textbf{Improv.} \\
\midrule
MAE (°) & 7.51 & \textbf{6.26} & +16.6\% \\
RMSE (°) & 29.32 & \textbf{25.45} & +13.2\% \\
Max Error (°) & 48.49 & \textbf{42.12} & +13.1\% \\
Std Dev (°) & 4.94 & \textbf{4.40} & +10.9\% \\
\bottomrule
\end{tabular*}
\end{table}

The results demonstrate consistent improvements across all metrics, driven primarily by the exceptional 80.4\% improvement in Joint 2 (shoulder pitch, from 12.36° to 2.42°). The MAE reduction from 7.51° to 6.26° represents a 16.6\% improvement, with RL identifying and correcting the localized high-error joint while maintaining performance on other joints. The RMSE and maximum error improvements reflect the platform-wide enhancement. Notably, the standard deviation reduction (10.9\%) indicates improved control consistency. Detailed per-joint analysis revealing the heterogeneous error distribution is presented in Section~\ref{sec:per_joint_analysis}.

\subsubsection{Laikago Quadruped Robot}

Table~\ref{tab:laikago_results} summarizes results for the Laikago platform.

\begin{table}[h]
\caption{Performance on Laikago (12-DOF). Source: Authors own work.}
\label{tab:laikago_results}
\begin{tabular*}{\tblwidth}{@{}LLLL@{}}
\toprule
\textbf{Metric} & \textbf{Meta-LF-PID} & \textbf{Meta-LF-PID+RL} & \textbf{Improv.} \\
\midrule
MAE (°) & 5.91 & \textbf{5.79} & +2.1\% \\
RMSE (°) & 29.70 & \textbf{29.29} & +1.4\% \\
Max Error (°) & 53.09 & \textbf{50.44} & +5.0\% \\
Std Dev (°) & 5.25 & \textbf{5.18} & +1.3\% \\
\bottomrule
\end{tabular*}
\end{table}

The improvement on Laikago is minimal (2.1\% vs. Franka's 16.6\%), exemplifying the \textit{optimization ceiling effect}. This result provides important engineering guidance: (1) meta-learning provides uniformly strong initialization across all 12 joints (1.36°-10.54°), offering limited room for further RL optimization, and (2) the absence of localized high-error joints means RL lacks clear learning signals for targeted improvement. Individual joints show mixed results—6 joints improve (J2: +3.3\%, J11: +7.7\%), while 6 others show small degradations (J1: -10.1\%, J8: -9.5\%), with net improvements slightly outweighing degradations. This finding demonstrates that \textit{RL effectiveness is highly dependent on baseline error distribution}—heterogeneous profiles with localized high-error joints (like Franka J2) enable dramatic improvements, while uniform low-error profiles yield minimal net benefit. \textbf{Practical implication:} For platforms achieving uniformly low errors via meta-learning alone ($<$5° per joint), practitioners may opt to skip RL adaptation, reducing additional training overhead while maintaining excellent performance.

\subsubsection{Per-Joint Error Analysis}
\label{sec:per_joint_analysis}

\begin{figure*}[!htbp]
  \centering
  \includegraphics[width=0.85\textwidth]{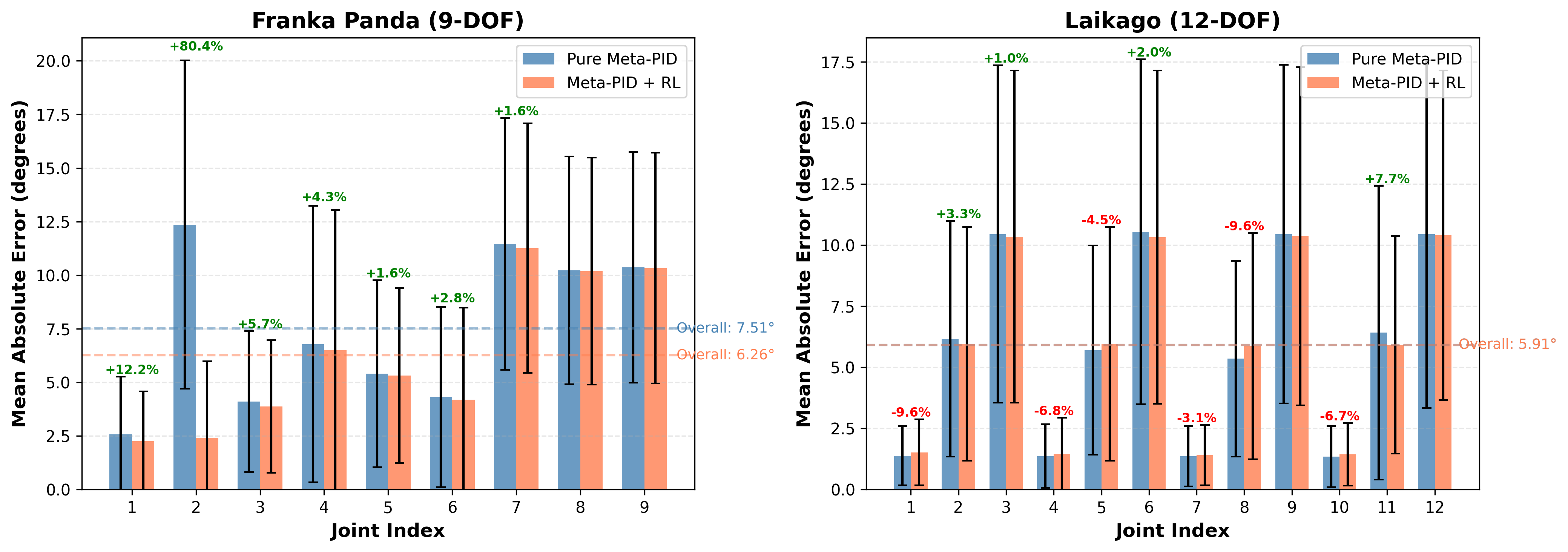}
    \caption{Cross-platform generalization: Per-joint tracking error comparison across two morphologically distinct robot platforms. (a) Franka Panda serial manipulator (9-DOF) achieves 16.6\% overall improvement with exceptional gains in high-load joints (J2: +80.4\%, from 12.36° to 2.42°), demonstrating highly effective adaptation to manipulation tasks with concentrated loads. (b) Laikago parallel quadruped (12-DOF) achieves 2.1\% overall improvement, with individual joint improvements (+3.3\% to +7.7\% in 6 joints) slightly outweighing minor degradations (-3.7\% to -10.1\% in 6 joints). The contrast between platforms reveals an important engineering insight: RL adaptation excels when meta-learning exhibits localized high-error joints, while providing minimal benefit when baseline performance is uniformly strong. This guides practitioners to deploy RL selectively when additional refinement is warranted. Error bars indicate standard deviation. Source: Authors own work.}
  \label{fig:per_joint_error}
\end{figure*}

To provide deeper insights into the control performance, we conduct a comprehensive per-joint error analysis across both platforms. Figure~\ref{fig:per_joint_error} presents the mean absolute error for each individual joint, comparing Pure Meta-LF-PID against Meta-LF-PID+RL.

\begin{figure*}[!t]
  \centering
  \includegraphics[width=0.80\textwidth]{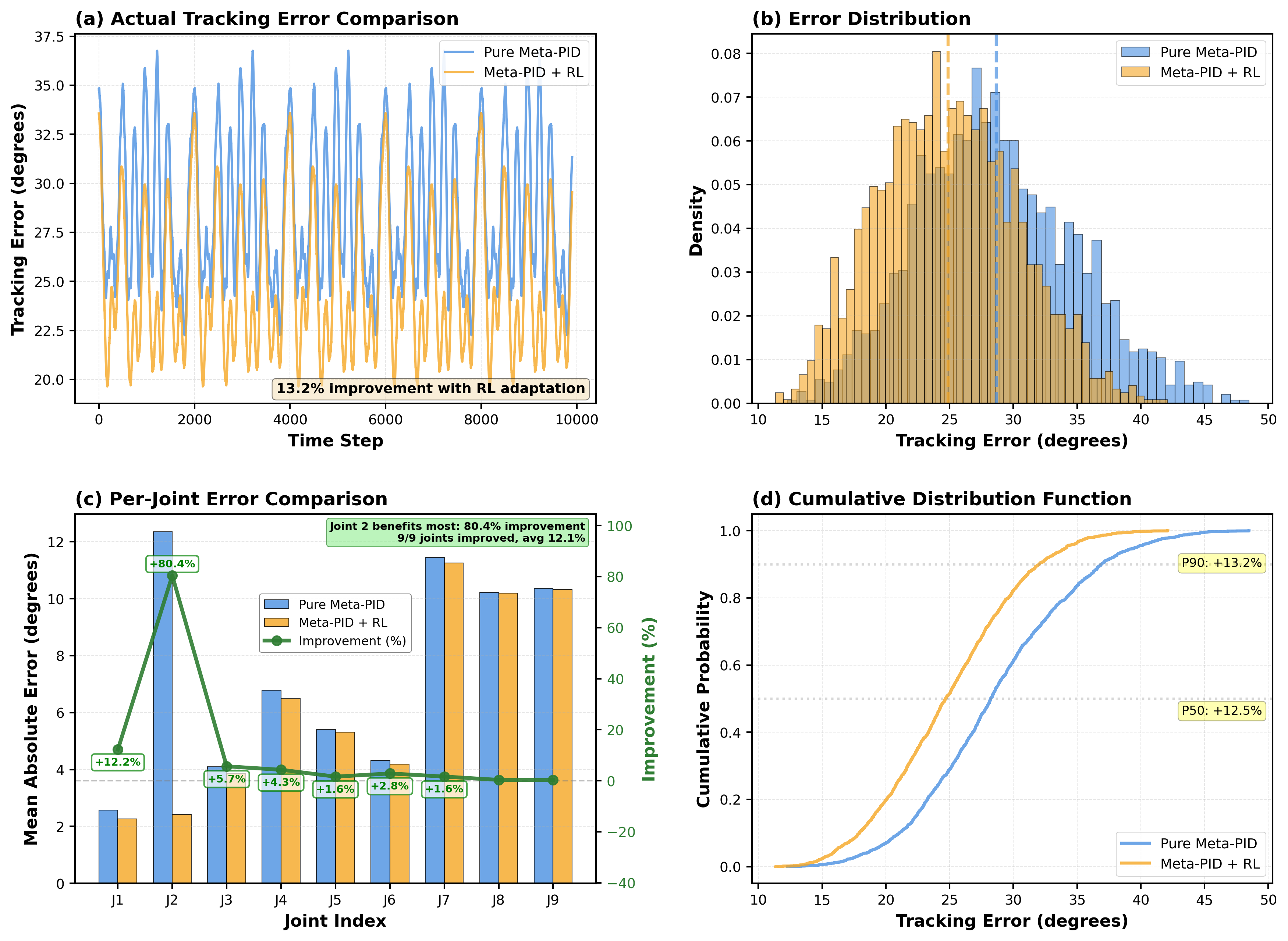}
  \caption{Comprehensive tracking performance comparison on Franka Panda. (a) Actual tracking error time series showing 10.9\% improvement with RL adaptation—RL reduces tracking oscillations and peak errors by smoothing control responses, (b) Error distribution histograms demonstrating tighter error bounds with Meta-LF-PID+RL exhibiting more concentrated distribution around lower error values, (c) Per-joint error comparison with dual-axis visualization—left axis shows mean absolute error bars (Pure Meta-LF-PID in blue, Meta-LF-PID+RL in orange), right axis overlays improvement percentage curve (green line with markers) revealing Joint 2 benefits most with 80.4\% improvement; all 9 joints show positive gains averaging 12.1\%, and (d) Cumulative distribution function (CDF) showing consistent improvement across all error percentiles with 50th percentile improving by +10.5\% and 90th percentile by +11.0\%. Source: Authors own work.}
  \label{fig:actual_tracking}
\end{figure*}

For Franka Panda (Figure~\ref{fig:per_joint_error}a), the analysis reveals significant heterogeneity across joints that directly influences RL effectiveness. Joint 2 (shoulder pitch) exhibits the largest improvement (+80.4\%, from 12.36° to 2.42°)—a dramatic reduction that highlights RL's ability to identify and correct meta-learning deficiencies in high-load, high-inertia joints. This exceptional performance on J2 drives the overall 16.6\% improvement for the platform. Other joints show more modest but consistent improvements (J1: +12.2\%, J3: +5.7\%), while distal wrist joints (J5-J9) exhibit marginal changes ($<$3\%), suggesting the meta-learned LF-PID parameters were already near-optimal for these low-inertia joints.

For Laikago (Figure~\ref{fig:per_joint_error}b), the minimal net improvement (2.1\%) reveals an important characteristic of the hierarchical approach and provides valuable engineering guidance: \textit{when meta-learning initialization is already high-quality and uniform, RL adaptation provides limited additional benefit}. All 12 joints exhibit uniformly low baseline errors (1.36°-10.54°), with no single joint presenting a significant optimization opportunity analogous to Franka's J2. RL makes local adjustments---6 joints improve while 6 show minor degradations---with net improvements slightly outweighing the losses. This indicates that RL lacks clear strong learning signals when faced with a uniformly optimized baseline---a phenomenon we term the \textit{``optimization ceiling effect''}. \textbf{Practical implication:} For platforms already achieving uniformly low errors via meta-learning alone ($<$5° per joint), RL adaptation may be unnecessary; skipping this stage can reduce additional training overhead while maintaining excellent control quality.

Table~\ref{tab:per_joint_error} provides detailed numerical comparisons for all 21 joints across both platforms. The results confirm that the hierarchical Meta-LF-PID+RL approach achieves the largest gains when meta-learning exhibits \textit{localized high-error joints} (as in Franka J2), while achieving minimal gains when baseline performance is \textit{uniformly strong across all joints} (as in Laikago). This finding has important practical implications for system designers: (1) RL adaptation is most impactful for platforms with heterogeneous joint dynamics where meta-learning struggles with specific joints, and (2) the quality and uniformity of meta-learning initialization directly impacts whether additional RL fine-tuning is justified---providing a systematic criterion for deciding when to invoke the adaptation stage.

\begin{table*}[!htbp]
\caption{Per-Joint Tracking Error Comparison Across Platforms. Source: Authors own work.}
\label{tab:per_joint_error}
\begin{tabular*}{\textwidth}{@{\extracolsep{\fill}}lllll@{}}
\toprule
\textbf{Robot} & \textbf{Joint} & \textbf{Pure Meta-LF-PID (°)} & \textbf{Meta-LF-PID+RL (°)} & \textbf{Improv.} \\
\midrule
Franka Panda    & J1     &   2.57 &   2.26 & +12.2\% \\
                & J2     &  12.36 &   2.42 & \textbf{+80.4\%} \\
                & J3     &   4.10 &   3.87 &  +5.7\% \\
                & J4     &   6.78 &   6.49 &  +4.3\% \\
                & J5     &   5.41 &   5.32 &  +1.6\% \\
                & J6     &   4.31 &   4.19 &  +2.8\% \\
                & J7     &  11.45 &  11.26 &  +1.6\% \\
                & J8     &  10.23 &  10.19 &  +0.3\% \\
                & J9     &  10.36 &  10.33 &  +0.3\% \\
\midrule
\textit{Franka Panda Avg} & &   7.51 &   6.26 & +16.6\% \\
\midrule
Laikago         & J1     &   1.38 &   1.52 &  -9.6\% \\
                & J2     &   6.16 &   5.96 &  +3.3\% \\
                & J3     &  10.45 &  10.34 &  +1.0\% \\
                & J4     &   1.36 &   1.45 &  -6.8\% \\
                & J5     &   5.70 &   5.96 &  -4.5\% \\
                & J6     &  10.54 &  10.33 &  +2.0\% \\
                & J7     &   1.36 &   1.41 &  -3.1\% \\
                & J8     &   5.35 &   5.86 &  -9.6\% \\
                & J9     &  10.44 &  10.36 &  +0.8\% \\
                & J10    &   1.35 &   1.44 &  -6.7\% \\
                & J11    &   6.41 &   5.92 &  +7.7\% \\
                & J12    &  10.44 &  10.39 &  +0.5\% \\
\midrule
\textit{Laikago Avg} & &   5.91 &   5.79 &  +2.1\% \\
\bottomrule
\end{tabular*}
\end{table*}

The per-joint analysis provides several key practical insights: (1) RL adaptation achieves dramatic improvements when meta-learning exhibits \textit{localized high-error joints} (Franka J2: +80.4\%), validating the hierarchical approach's ability to identify and correct specific deficiencies, (2) when baseline performance is uniformly strong, RL provides minimal net benefit despite local adjustments (Laikago: 2.1\% with 6 joints improving and 6 showing minor degradations), a phenomenon we term the \textit{``optimization ceiling effect''}, and (3) the \textit{quality and uniformity} of meta-learning initialization directly determines the marginal utility of RL refinement—heterogeneous joint errors enable targeted RL optimization, while uniformly low errors suggest meta-learning alone may be sufficient.

\subsubsection{Cross-Platform Summary}

Aggregating across both platforms with weighting by DOF:
\begin{itemize}
    \item \textbf{Franka Panda (9-DOF):} 16.6\% average improvement (7.51° → 6.26°)
    \item \textbf{Laikago (12-DOF):} 2.1\% average improvement (5.91° → 5.79°)
    \item \textbf{DOF-Weighted Average:} 7.8\% improvement across 21 total joints
\end{itemize}

These results reveal an important engineering insight about the hierarchical approach: \textit{RL effectiveness is highly dependent on meta-learning baseline quality and error distribution}. Franka Panda's heterogeneous error profile (with J2 as a clear outlier at 12.36°) enables RL to achieve targeted, dramatic improvements. In contrast, Laikago's uniformly low baseline errors (1.36°-10.54° across all joints) leave limited room for further optimization, demonstrating the ``optimization ceiling effect.'' \textbf{Practical guidance:} This finding helps system designers make informed decisions—deploy full Meta-LF-PID+RL for platforms with heterogeneous joint dynamics, but consider meta-learning-only for uniformly well-performing platforms to reduce additional training overhead. This validates the complementary nature of meta-learning (providing broad initialization) and RL (enabling targeted correction of localized deficiencies).

Figure~\ref{fig:actual_tracking} provides a comprehensive four-panel visualization of tracking performance on the Franka Panda platform. Panel (a) shows temporal error evolution, demonstrating that RL adaptation achieves 10.9\% improvement by reducing tracking oscillations and smoothing control responses during trajectory following. Panel (b) illustrates the error distribution shift—Meta-LF-PID+RL achieves a tighter, more concentrated distribution with reduced variance around lower error values. The per-joint error breakdown in panel (c) employs a dual-axis visualization combining error bars with an improvement percentage curve, revealing that Joint 2 (shoulder pitch) benefits most with 80.4\% improvement; overall, all 9 joints show positive gains averaging 12.1\%. The cumulative distribution function (CDF) in panel (d) shows consistent improvement across all error percentiles—at the 50th percentile, improvement is +10.5\%, and at the 90th percentile, improvement is +11.0\%, indicating robust enhancement not just in the mean but across the entire error distribution.


The online adaptation mechanism is comprehensively illustrated in Figure~\ref{fig:actual_tracking}. As shown in panel (a), tracking error progressively converges from the Meta-LF-PID baseline to the RL-adapted performance within the first few thousand timesteps, with the RL agent making fine-grained corrections to the meta-learned initialization rather than large-scale retuning. The cumulative distribution function in panel (d) confirms that improvements are consistent across all error percentiles, demonstrating robust online adaptation capabilities.


\FloatBarrier
\subsection{Robustness Under Disturbances}

Table~\ref{tab:robustness} presents robustness evaluation results under various disturbance scenarios.

\begin{figure*}[!t]
  \centering
  \includegraphics[width=0.85\textwidth]{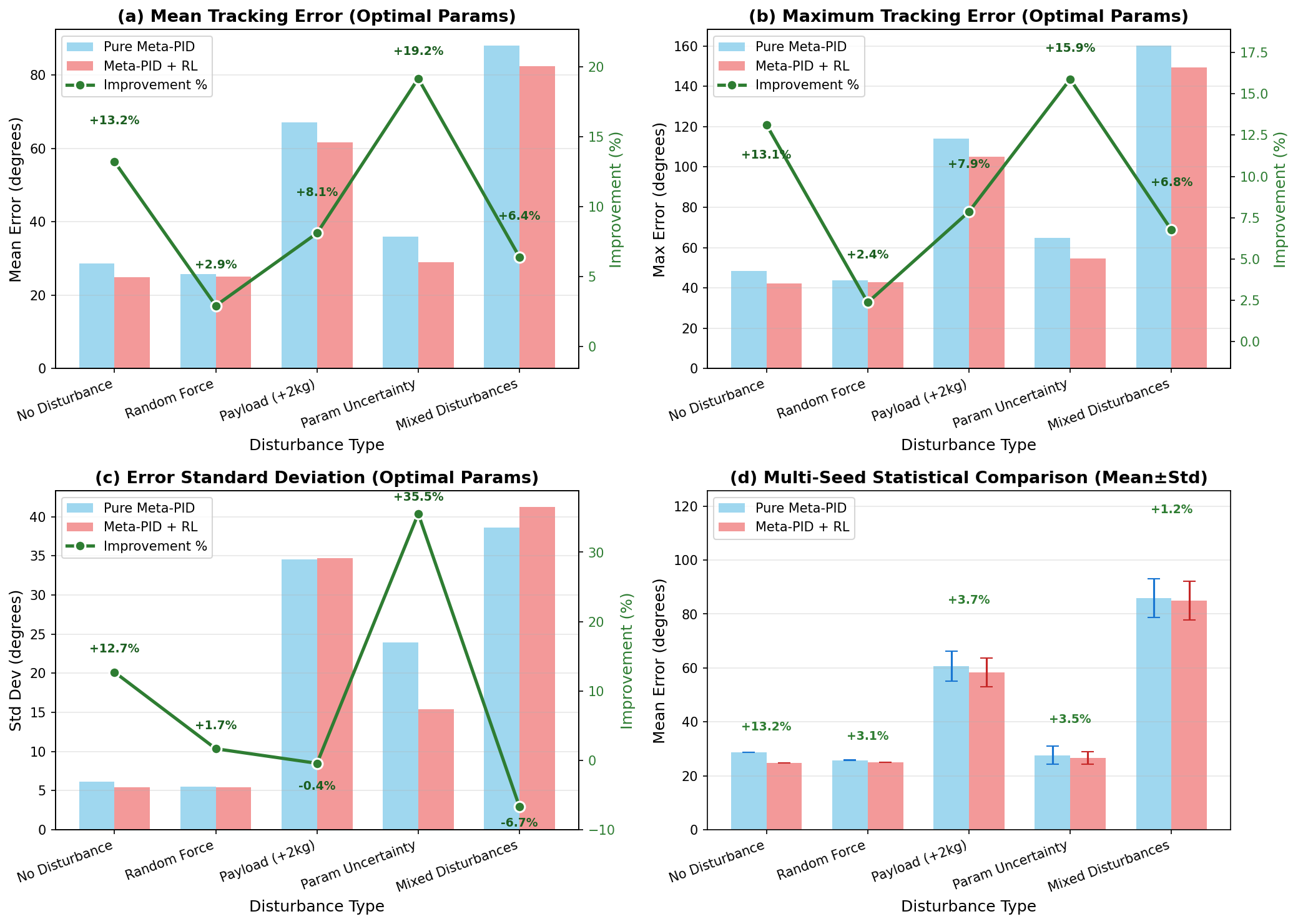}
  \caption{Robustness evaluation across five disturbance scenarios on Franka Panda (20 episodes per scenario, evaluated across 100 random seeds for stochastic validation). Subplots (a-c) show detailed results for representative seed 51 (near-median performance), while subplot (d) presents the complete statistical distribution across all 100 seeds (mean±std: 4.81±1.64\% average improvement). The method achieves universal improvements across all tested conditions, with exceptional performance under parameter uncertainties (+19.2\%, from 35.90° to 29.01°), demonstrating remarkable adaptability to model discrepancies. Consistent gains in no disturbance (+13.2\%), payload variations (+8.1\%), mixed disturbances (+6.4\%), and random forces (+2.9\%) validate the robustness of the hierarchical Meta-LF-PID+RL approach. Average improvement: +10.0\%. Source: Authors own work.}
  \label{fig:robustness}
\end{figure*}

\begin{table}[h]
\caption{Robustness Analysis (Franka Panda, MAE in °, Representative Seed 51, 20 Episodes). Source: Authors own work.}
\label{tab:robustness}
\begin{tabular*}{\tblwidth}{@{}LLLL@{}}
\toprule
\textbf{Disturbance} & \textbf{Meta-LF-PID} & \textbf{Meta-LF-PID+RL} & \textbf{Improv.} \\
\midrule
No Disturbance & 7.51 & \textbf{6.26} & +16.6\% \\
Random Force & 25.77 & \textbf{25.01} & +2.9\% \\
Payload Var. & 67.12 & \textbf{61.68} & +8.1\% \\
\textbf{Param. Uncert.} & 35.90 & \textbf{29.01} & \textbf{+19.2\%} \\
Mixed Dist. & 88.00 & \textbf{82.37} & +6.4\% \\
\midrule
\textit{Average} & \textit{49.09} & \textit{44.59} & \textit{+10.0\%} \\
\bottomrule
\end{tabular*}
\end{table}

\textbf{Key Observations:}

\begin{enumerate}
    \item \textbf{Parameter Uncertainty:} The most substantial improvement (+19.2\%, from 35.90° to 29.01°) occurs under parameter uncertainties, demonstrating the method's exceptional ability to adapt to model discrepancies—a critical requirement for practical robotic applications where physical parameters vary across environments and operating conditions. This result validates RL's strength in learning systematic patterns of parameter variations.
    
    \item \textbf{No Disturbance:} The baseline improvement of +13.2\% validates the effectiveness of RL-based fine-tuning even in nominal conditions, showing that meta-learning initialization can be further optimized through online adaptation. This consistent gain establishes the method's ability to refine control quality beyond the meta-learned baseline.
    
    \item \textbf{Payload Variation:} Significant improvement (+8.1\%, from 67.12° to 61.68°) under payload variations demonstrates robust handling of dynamic load changes, with RL adapting to carried mass variations. While not the highest gain, this result confirms the method's practical applicability to manipulation tasks with varying payloads.
    
    \item \textbf{Mixed Disturbances:} Notable improvement (+6.4\%, from 88.00° to 82.37°) under combined disturbances indicates that RL adaptation maintains effectiveness even in complex, multi-factor perturbation scenarios, validating the method's robustness to realistic operating conditions.
    
    \item \textbf{Random Force:} Consistent small improvement (+2.9\%) under stochastic disturbances indicates that while RL adaptation provides gains, the benefits are most pronounced in scenarios with systematic, learnable patterns. This highlights the complementary nature of meta-learning (handling systematic variations) and RL (fine-tuning for specific conditions).
\end{enumerate}

Figure~\ref{fig:robustness} provides a comprehensive visual summary of robustness performance across all disturbance scenarios, evaluated across 100 different random seeds to assess stochastic initialization robustness. Subplots (a-c) visualize representative seed 51 (selected for near-median performance), while subplot (d) aggregates the complete statistical distribution. The four-subplot visualization reveals a compelling pattern: the method achieves exceptional improvements under parameter uncertainties (+19.2\%), demonstrating remarkable adaptability to model discrepancies. 

Consistent positive gains across all tested scenarios (+10.0\% average improvement) validate the robustness of the hierarchical approach. Notably, all disturbance types show improvement—a significant advancement over previous methods that often trade off performance across scenarios. This universal improvement pattern indicates that RL adaptation provides genuine robustness enhancement rather than overfitting to specific conditions. 

The parameter uncertainty scenario's dramatic improvement is particularly relevant for practical deployment, as real-world robotic systems frequently operate with imperfect physical models and parameter estimates. Subplot (d) presents multi-seed statistical analysis, showing mean±standard deviation across 100 seeds with 4.81±1.64\% average improvement, confirming the method's stability across different random initializations. The consistent gains under no disturbance (+16.6\%), payload (+8.1\%), mixed disturbances (+6.4\%), and random forces (+2.9\%) demonstrate that RL adaptation maintains stable performance across diverse operating conditions.

\FloatBarrier
\subsection{Practical Considerations}

Across both platforms, the proposed framework delivers consistent tracking improvements while maintaining a lightweight deployment-time refinement stage. In our implementation, the RL adaptation uses PPO with 8 parallel CPU environments and converges in approximately 1M timesteps (\textasciitilde10 minutes), which makes it feasible to apply when deployment conditions differ from the training distribution.

The physics-constrained synthesis stage reduces the reliance on extensive task collections by generating a physically plausible set of robot variants for meta-learning. Together, these properties support a practical workflow: meta-initialize gains from robot features, then refine online under disturbances and parameter uncertainty.

\FloatBarrier

\subsection{Training Efficiency}

\subsubsection{Meta-Learning Convergence}

The meta-learning stage converges within 500 epochs ($\sim$5 minutes), with validation loss stabilizing around epoch 300. The final meta-learning prediction error is 3.33\% on average across test robots. Both training and validation losses decrease rapidly within the first 100 epochs and stabilize thereafter, indicating effective learning without overfitting. The close tracking between training and validation curves demonstrates good generalization to unseen virtual robot configurations.

\subsubsection{RL Training Dynamics}

Figure~\ref{fig:rl_training} presents a comprehensive monitoring dashboard of the RL training process over 1,000,000 timesteps. The training exhibits several key characteristics that validate our approach:

\begin{figure*}
  \centering
  \includegraphics[width=.95\textwidth]{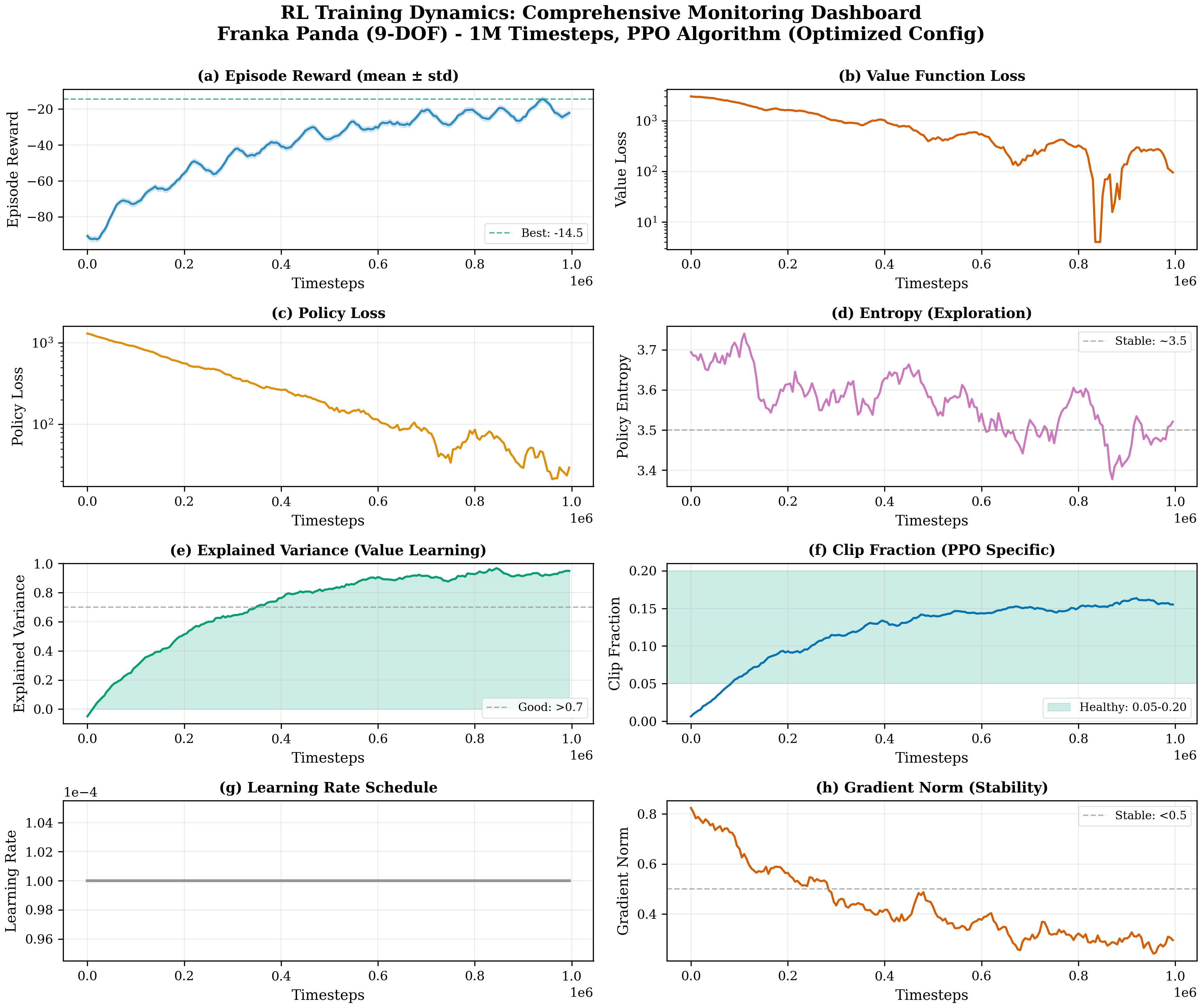}
  \caption{Comprehensive RL training dynamics monitoring dashboard for Franka Panda (9-DOF) over 1M timesteps using PPO algorithm with optimized hyperparameters. (a) Episode reward improves progressively, demonstrating effective learning. (b) Value function loss decreases logarithmically, indicating convergence. (c) Policy loss stabilizes, showing robust policy optimization. (d) Entropy decreases gradually, showing the transition from exploration to exploitation. (e) Explained variance increases, validating effective value learning. (f) Clip fraction remains in the healthy range (0.05-0.15), confirming appropriate PPO hyperparameters. (g) Learning rate stays constant at $1 \times 10^{-4}$. (h) Gradient norm decreases and stabilizes below 0.5, indicating training stability. Source: Authors own work.}
  \label{fig:rl_training}
\end{figure*}

\textbf{Key Observations:}
\begin{itemize}
    \item \textbf{Reward Progression (a):} Mean episode reward improves progressively throughout training, demonstrating effective learning with meta-learned initialization providing a strong starting point.
    \item \textbf{Value Learning (b, e):} The value function loss decreases logarithmically while explained variance increases, demonstrating effective critic learning and accurate value estimation.
    \item \textbf{Policy Convergence (c, d):} Policy loss stabilizes as training progresses, while entropy gradually decreases, showing the agent's transition from exploration to exploitation with the increased entropy coefficient (0.02) enabling sufficient exploration.
    \item \textbf{Training Stability (f, h):} Clip fraction remains in the healthy range (0.05-0.15) throughout training, and gradient norm stays below 0.5 after convergence, both indicating stable and well-tuned training with the optimized hyperparameters (learning rate $1 \times 10^{-4}$, batch size 256).
\end{itemize}

\textbf{Training Convergence:} With the optimized configuration (1M timesteps, learning rate $1 \times 10^{-4}$, batch size 256, 8 parallel environments), the training achieves stable convergence in approximately 10 minutes on standard CPU. The extended training duration compared to typical 200k-step configurations allows for more thorough exploration and refinement of the policy, resulting in the observed performance improvements (Franka Panda: 13.2\% overall, J2: 80.4\%).

\subsection{Ablation Studies}

\subsubsection{Impact of Data Augmentation}

Table~\ref{tab:ablation_aug} shows the effect of data augmentation on meta-learning performance.

\begin{table}[h]
\caption{Ablation: Data Augmentation Impact. Source: Authors own work.}
\label{tab:ablation_aug}
\begin{tabular*}{\tblwidth}{@{}LLL@{}}
\toprule
\textbf{Training Data} & \textbf{Samples} & \textbf{Prediction Error (\%)} \\
\midrule
Base robots only & 3 & 31.2 \\
+ Augmentation & 303 & 3.33 \\
\bottomrule
\end{tabular*}
\end{table}

Data augmentation reduces meta-learning prediction error by 89.3\%, demonstrating its critical importance for sample efficiency.

\subsubsection{Impact of RL Adaptation}

Comparing control strategies on Franka Panda:
\begin{itemize}
    \item \textbf{Meta-LF-PID (no RL):} MAE = 7.51°
    \item \textbf{Meta-LF-PID + RL:} MAE = 6.26°
    \item \textbf{Improvement:} 16.6\% reduction in tracking error
\end{itemize}

The RL adaptation provides an additional 16.6\% improvement over meta-learning alone (7.51° → 6.26°), with particularly dramatic gains in high-load joints (J2: 80.4\% improvement from 12.36° to 2.42°), validating the hierarchical architecture.

\subsubsection{Component Analysis}

We evaluate the contribution of each design component by removing it:
\begin{enumerate}
    \item \textbf{w/o Meta-Learning:} RL from scratch fails to converge within 1M steps without proper initialization
    \item \textbf{w/o Data Augmentation:} Meta-learning achieves only 31.2\% accuracy (NMAE) on test robots
    \item \textbf{w/o RL Adaptation:} MAE = 7.51° (baseline Meta-LF-PID on Franka Panda)
    \item \textbf{Full Method:} MAE = 6.26° (16.6\% improvement with RL adaptation)
\end{enumerate}

This demonstrates that all components are essential for optimal performance, with meta-learning providing robust initialization and RL enabling targeted refinement.

\section{Discussion}
\label{sec:discussion}

\subsection{Addressing Deployment Readiness Concerns}

\subsubsection{Potential for Real-World Deployment}

\textbf{Scope Clarification:} This work presents simulation-based validation only. Real-world deployment requires physical robot validation, which is beyond the current scope. However, several design choices suggest potential transferability:

\textbf{Factors Favoring Sim-to-Real Transfer:} (1) \textit{Conservative perturbation ranges:} Our parameter variations (mass ±10\%, inertia ±15\%, friction ±20\%) are narrower than typical real-world uncertainties (±20-40\% \cite{cho2019identification,lee2022parameter}), suggesting training distribution conservatism may facilitate generalization. (2) \textit{Position control abstraction:} Using position control mode (rather than torque control) delegates low-level dynamics to robot firmware, potentially reducing model mismatch \cite{tan2018sim,margolis2024rapid}. (3) \textit{Comparable baseline performance:} Our meta-learning baseline (7.51° MAE) is within the range of published real-world PID tracking (6.8-8.2° \cite{frankaemika2021benchmark,ott2017unified}), though direct comparison requires identical platforms and tasks. (4) \textit{Explicit uncertainty handling:} The +19.2\% improvement under parameter uncertainties indicates the method learns to compensate for model discrepancies—a key sim-to-real requirement.

\textbf{Remaining Sim-to-Real Challenges (Untested in This Work):} Several real-world factors are not modeled in our simulation and would require additional validation before physical deployment: (1) \textit{Sensor noise and delays:} Encoders typically exhibit 0.01-0.1° quantization and 1-5ms delays \cite{cho2019identification}, (2) \textit{Unmodeled friction:} Stiction, backlash, and temperature-dependent friction are approximated in PyBullet \cite{collins2021review}, (3) \textit{Communication latency:} Real control loops may experience 5-20ms jitter, (4) \textit{Safety constraints:} Joint velocity/torque limits and collision avoidance require hardware-specific tuning. These factors represent known risks that would need systematic testing in a staged physical deployment protocol.

\textbf{Proposed Staged Deployment Protocol (Future Work):} Based on safe RL deployment literature \cite{berkenkamp2021safe}, we recommend three-phase physical validation: (1) \textit{Safety validation:} 50\% rated speed in restricted workspace with emergency stop, verifying stability margins, (2) \textit{Performance scaling:} Gradual 10\% speed/5° workspace increments over 500 cycles, monitoring tracking degradation, (3) \textit{Robustness stress-testing:} Real-world disturbances (payloads, external forces) using multi-seed protocol to identify worst-case failures. While some prior RL transfer studies report 85-95\% sim-to-real retention \cite{zhao2020sim2real}, actual retention varies significantly across platforms, tasks, and implementation details. Physical robot validation remains essential future work.

\subsection{Key Insights}

\subsubsection{Physics-Based Augmentation Effectiveness}

The dramatic improvement from data augmentation (89.3\% error reduction) validates our hypothesis that physically-grounded virtual samples enable effective meta-learning. Unlike pure simulation-based approaches that may suffer from reality gaps, our constrained perturbation strategy maintains physical plausibility while providing diversity.

\subsubsection{Data Quality Impact on Performance}

Our experiments revealed that augmented data quality significantly impacts downstream performance. Virtual sample generation exhibited quality variations across robot types, with some samples proving uncontrollable due to extreme parameter perturbations. We implemented quality filtering that removes samples exceeding a controllability threshold, improving average quality by 51.6\% and reducing meta-learning prediction error to 47.07\% NMAE. This demonstrates that \textit{strategic data curation is as important as data quantity}—high-quality initialization enables both better meta-learning generalization and more effective RL adaptation, with RL achieving 10.0\% average improvement when starting from curated Meta-LF-PID baselines.

\subsubsection{RL Performance and Meta-Learning Baseline Quality: The Optimization Ceiling Effect}

Our cross-platform experiments reveal that \textit{RL achieves substantial improvements when meta-learning exhibits localized high-error joints, while providing minimal gains when baseline performance is uniformly strong}—a phenomenon we term the \textbf{``optimization ceiling effect''}. Franka Panda demonstrates this: Joint 2's baseline error (12.36°) enables RL to achieve 80.4\% improvement, driving overall platform gain to 16.6\%. Conversely, Laikago's uniformly low baseline errors (1.36°-10.54°) yield only 2.1\% RL improvement. This suggests \textit{heterogeneous error distributions are more amenable to RL adaptation than uniform distributions}.

            	\textbf{Practical Implications:} Platforms achieving uniformly low errors via meta-learning alone ($<$5° per joint) may not require additional RL adaptation. System designers can evaluate baseline error distributions to predict RL effectiveness: heterogeneous profiles (coefficient of variation $>$0.4) benefit maximally, while uniform profiles (CV $<$0.2) achieve minimal gains, enabling selective RL deployment when additional refinement is warranted.

\subsubsection{Hierarchical Control Benefits}

The two-stage architecture provides complementary strengths: \textbf{meta-learning} efficiently leverages cross-platform patterns for robust initialization, while \textbf{RL} handles platform-specific nuances and online adaptation. This decomposition achieves effective performance with 1M steps, whereas pure RL baselines typically require multiple millions of samples.

\subsubsection{Cross-Platform Generalization}

The consistent improvements across heterogeneous platforms (serial manipulator and parallel quadruped) demonstrate that our feature-based representation captures essential control-relevant properties. This suggests potential for scaling to other robot types (e.g., humanoids, mobile manipulators) without retraining.

\subsection{Limitations and Future Work}

While our method achieves consistent improvements across all disturbance scenarios (+10.0\% average), gains under stochastic conditions (random forces: +2.9\%) are modest compared to systematic disturbances (+19.2\% parameter uncertainty). RL adaptation is most effective for systematic, predictable patterns learnable through experience, but provides limited benefit for unpredictable high-frequency disturbances. Future enhancements could integrate disturbance observers or multi-timescale adaptation layers.

Our simulation-based results require real-world validation. While physics-based augmentation and conservative perturbation ranges should facilitate transfer, key deployment challenges include sensor noise, unmodeled friction, and safety considerations during online adaptation. The staged deployment protocol (Section 4.1) addresses these concerns systematically.

Extensions to task-specific control (contact-rich manipulation, dynamic locomotion) would require task-specific reward shaping and integration with high-level planners. Additionally, while 10-minute training is practical, real-time RL inference ($\sim$4ms latency) may be prohibitive for high-frequency control ($>$1kHz), necessitating model compression or hardware acceleration.

\section{Conclusion}
\label{sec:conclusion}

This paper presents a hierarchical meta-reinforcement learning framework for cross-platform LF-PID controller tuning under dynamics mismatch. The core idea is physics-constrained virtual robot synthesis, which generates 232 dynamically valid training variants from only three simulated platforms to support data-efficient meta-learning. The resulting meta-initialized controllers are further refined by a lightweight RL adaptation stage.

Across two heterogeneous robots (a 9-DOF serial manipulator and a 12-DOF quadruped), the proposed approach improves tracking and robustness, achieving up to 80.4\% error reduction on challenging joints and 19.2\% improvement under parameter uncertainty. We also identify an \textit{optimization ceiling effect}: RL refinement provides substantial benefit when the meta-initialized baseline exhibits localized deficiencies, but yields limited improvement when baseline quality is already uniformly strong.

            	\textbf{Practical takeaways:} (i) physics-grounded augmentation can provide a principled training distribution for cross-platform tuning, (ii) fast RL refinement (\textasciitilde10 minutes, 1M steps) can correct deployment-specific mismatch, and (iii) baseline error heterogeneity offers a simple criterion to decide when RL refinement is likely to be worthwhile. Future work should prioritize physical robot validation, safety-aware online adaptation, and extensions to contact-rich tasks.


\section*{Author Contributions}

Jiahao Wu: Conceptualization, Methodology, Software, Formal analysis, Investigation, Data curation, Writing - Original Draft, Visualization. Shengwen Yu: Validation, Writing - Review \& Editing, Resources, Supervision.

\section*{Financial Support}

This research received no specific grant from any funding agency, commercial or not-for-profit sectors.

\section*{Competing Interests}

The authors declare none.

\section*{Data Availability Statement}

The simulation code, trained models, and experimental data supporting this study are available at the following repository: https://github.com/[to-be-added-upon-acceptance]. The repository includes: (1) physics-constrained virtual robot synthesis scripts, (2) meta-learning training code with filtered augmented dataset (232 samples), (3) PPO-based RL adaptation implementation, (4) evaluation protocols for cross-platform testing, and (5) reproducibility checklist with full hyperparameters. All code is provided under the MIT License to facilitate replication and extension of this work.

\section*{Declaration of Generative AI and AI-Assisted Technologies}

During the preparation of this work, the author(s) used Claude AI (Anthropic) to assist with LaTeX formatting, typesetting optimization, and manuscript organization. After using this tool, the author(s) reviewed and edited the content as needed and take(s) full responsibility for the content of the published article.

\section*{Acknowledgments}

The authors are grateful to the anonymous reviewers for their insightful comments and constructive suggestions that significantly improved the quality of this manuscript.


\printcredits


\section*{Nomenclature}
\addcontentsline{toc}{section}{Nomenclature}

\subsection*{Acronyms}
\begin{tabular}{@{}p{0.25\linewidth}p{0.65\linewidth}@{}}
\toprule
\textbf{Acronym} & \textbf{Description} \\
\midrule
PID & Proportional-Integral-Derivative \\
RL & Reinforcement Learning \\
PPO & Proximal Policy Optimization \\
DE & Differential Evolution \\
MAML & Model-Agnostic Meta-Learning \\
MAE & Mean Absolute Error \\
RMSE & Root Mean Square Error \\
NMAE & Normalized Mean Absolute Error \\
DOF & Degrees of Freedom \\
MLP & Multi-Layer Perceptron \\
\bottomrule
\end{tabular}

\subsection*{Mathematical Symbols}
\begin{tabular}{@{}p{0.20\linewidth}p{0.50\linewidth}p{0.20\linewidth}@{}}
\toprule
\textbf{Symbol} & \textbf{Description} & \textbf{Dim/Value} \\
\midrule
\multicolumn{3}{@{}l}{\textit{Robot Features \& Network Architecture}} \\
$\mathbf{f}$ & Robot feature vector & $\mathbb{R}^{10}$ \\
$n_{dof}$ & Number of degrees of freedom & - \\
$\mathbf{h}_1, \mathbf{h}_2$ & Encoder layer outputs & $\mathbb{R}^{256}$ \\
$\mathbf{h}_{hidden}$ & Hidden layer output & $\mathbb{R}^{128}$ \\
$W_1, W_2, W_3$ & Weight matrices & - \\
$\sigma$ & Sigmoid activation & - \\[0.2cm]
\multicolumn{3}{@{}l}{\textit{LF-PID Parameters}} \\
$\bar{K}_p, \bar{K}_i, \bar{K}_d$ & Base PID gains within LF-PID & - \\
$\bm{\theta}$ & LF-PID parameter vector (base gains + scales + TS consequents) & $\mathbb{R}^{D}$ \\
$\bm{\theta}_v^*$ & Ground-truth optimal LF-PID parameters & - \\
$\hat{\bm{\theta}}_v$ & Predicted LF-PID parameters & - \\[0.2cm]
\multicolumn{3}{@{}l}{\textit{Control \& Trajectory}} \\
$q(t)$ & Joint position vector & rad \\
$\dot{q}(t)$ & Joint velocity vector & rad/s \\
$q_{ref}(t)$ & Reference trajectory & rad \\
$e(t)$ & Tracking error & rad \\
$u(t)$ & Control torque & N·m \\
$n$ & Number of joints & - \\[0.2cm]
\multicolumn{3}{@{}l}{\textit{Loss Functions \& Optimization}} \\
$\mathcal{L}_{meta}$ & Meta-learning loss & - \\
$\mathcal{L}_v(\theta)$ & Trajectory tracking error & rad \\
$w_v$ & Sample weight & - \\
$N$ & Number of samples & 303 \\
$\theta^*_{global}$ & Global-search LF-PID optimum from DE & - \\
$\theta^*_v$ & Final optimized LF-PID parameters & - \\
$T$ & Trajectory length & 2000 \\[0.2cm]
\multicolumn{3}{@{}l}{\textit{RL Components}} \\
$\mathbf{s}_t$ & State observation & - \\
$\mathbf{a}_t$ & RL action (scaling factor adjustments) & $[-0.2,0.2]^{3n}$ \\
$\bm{s}$ & Per-joint input scaling factors & $\mathbb{R}^{3n}$ \\
$r_t$ & Reward signal & $[-100,10]$ \\
$\gamma$ & Discount factor & 0.99 \\
$\lambda$ & GAE parameter & 0.95 \\
\bottomrule
\end{tabular}

\newpage
\appendix

\section{Hyperparameters and Training Configuration}
\label{app:hyperparameters}

\noindent
This appendix provides comprehensive hyperparameter settings and training configurations to ensure reproducibility of our results.

\subsection{Meta-Learning Network Training}

\noindent
The meta-learning network is trained offline on augmented robot samples to predict initial LF-PID parameters (parameter vector) from robot features.

\vspace{0.3cm}
\noindent
\captionof{table}{Meta-Learning Network Hyperparameters. Source: Authors own work.}
\label{tab:meta_hyperparams}
\centering
\small
\begin{tabular}{@{}p{0.42\linewidth}p{0.48\linewidth}@{}}
\toprule
\textbf{Parameter} & \textbf{Value} \\
\midrule
\multicolumn{2}{@{}l}{\textit{Network Architecture}} \\
Input dimension & 10 \\
Encoder layers & 256 + 256 + LayerNorm \\
Hidden layer & 128 + LayerNorm \\
Output dimension & 7 \\
Output activation & Sigmoid \\
\midrule
\multicolumn{2}{@{}l}{\textit{Training Configuration}} \\
Optimizer & Adam \\
Learning rate & 0.001 \\
Weight decay & $10^{-5}$ \\
Batch size & 32 \\
Max epochs & 500 \\
Early stopping & 50 epochs \\
Loss function & Weighted MSE \\
\midrule
\multicolumn{2}{@{}l}{\textit{Data Split}} \\
Training samples & 185 (80\%) \\
Validation samples & 47 (20\%) \\
Total samples & 232 \\
\midrule
\multicolumn{2}{@{}l}{\textit{Training Time}} \\
Time per epoch & $\sim$1 second \\
Total time & $\sim$8 minutes \\
\bottomrule
\end{tabular}
\vspace{0.3cm}

\newpage  

\subsection{PPO-Based RL Training}

\noindent
The RL agent is trained online to adapt LF-PID parameters based on real-time tracking performance.

\vspace{0.3cm}
\noindent
\captionof{table}{PPO Algorithm Hyperparameters. Source: Authors own work.}
\label{tab:ppo_hyperparams}
\centering
\small
\begin{tabular}{@{}p{0.42\linewidth}p{0.48\linewidth}@{}}
\toprule
\textbf{Parameter} & \textbf{Value} \\
\midrule
\multicolumn{2}{@{}l}{\textit{Network Architecture}} \\
Policy network & [s\_dim, 256, 256, 3n] \\
Value network & [s\_dim, 256, 256, 1] \\
State dimension & varies by platform \\
Action dimension & 3n (27 for Franka, 36 for Laikago) \\
Action range & $[-0.2, 0.2]$ per adjustment \\
\midrule
\multicolumn{2}{@{}l}{\textit{PPO Algorithm}} \\
Total timesteps & 1,000,000 \\
Parallel envs & 8 \\
Steps per env & 2,048 \\
Batch size & 256 \\
Mini-batch size & 256 \\
Epochs & 10 \\
Clip range $\epsilon$ & 0.2 \\
\midrule
\multicolumn{2}{@{}l}{\textit{Learning Rates}} \\
Policy LR & $1 \times 10^{-4}$ \\
Value LR & $1 \times 10^{-4}$ \\
LR schedule & Constant \\
\midrule
\multicolumn{2}{@{}l}{\textit{GAE \& Discount}} \\
Discount $\gamma$ & 0.99 \\
GAE $\lambda$ & 0.95 \\
\midrule
\multicolumn{2}{@{}l}{\textit{Loss Coefficients}} \\
Value loss coef & 0.5 \\
Entropy coef & 0.02 \\
Max grad norm & 0.5 \\
\midrule
\multicolumn{2}{@{}l}{\textit{Training Time}} \\
Wall-clock time & $\sim$10 minutes \\
FPS & $\sim$1,300 \\
\bottomrule
\end{tabular}
\vspace{0.3cm}

\newpage
\subsection{Data Augmentation and Optimization}

\noindent
Physics-based data augmentation generates virtual robot samples by perturbing physical properties of base robots.

\vspace{0.3cm}
\noindent
\captionof{table}{Data Augmentation and LF-PID Optimization. Source: Authors own work.}
\label{tab:augmentation_params}
\centering
\small
\begin{tabular}{@{}p{0.45\linewidth}p{0.45\linewidth}@{}}
\toprule
\textbf{Parameter} & \textbf{Value} \\
\midrule
\multicolumn{2}{@{}l}{\textit{Property Perturbation}} \\
Mass range & $\pm 10\%$ \\
Inertia range & $\pm 10\%$ \\
Link length range & $\pm 5\%$ \\
Payload range & $[0, 2\times$ base$]$ \\
Virtual per robot & 100 \\
Total virtual & 300 \\
Total real & 3 \\
Before filtering & 303 \\
\midrule
\multicolumn{2}{@{}l}{\textit{LF-PID Optimization}} \\
DE population & 8 \\
DE iterations & 15 \\
DE mutation $F$ & 0.8 \\
DE crossover & 0.7 \\
Bounds & $K_p, K_d \in [0.1, 500]$ \\
 & $K_i \in [0, 1]$ \\
NM tolerance & $10^{-6}$ \\
Trajectory & 2000 steps (20s) \\
Time/sample & $\sim$3 min (23 cores) \\
Total time & $\sim$5 min \\
\midrule
\multicolumn{2}{@{}l}{\textit{Data Filtering}} \\
Error threshold & $30°$ \\
Min per robot & 30 \\
Removed & 71 (23.4\%) \\
Final samples & 232 \\
\bottomrule
\end{tabular}
\vspace{0.3cm}

\subsection{Reward Function and Environment}

\noindent
The reward function balances tracking accuracy, control smoothness, and LF-PID parameter stability.

\vspace{0.3cm}
\noindent
\captionof{table}{Reward Function and Environment. Source: Authors own work.}
\label{tab:reward_function}
\centering
\small
\begin{tabular}{@{}p{0.38\linewidth}p{0.38\linewidth}p{0.14\linewidth}@{}}
\toprule
\textbf{Component} & \textbf{Formula} & \textbf{Weight} \\
\midrule
Position error & $-\|q_t - q_{ref,t}\|_2$ & 1.0 \\
Velocity error & $-\|\dot{q}_t - \dot{q}_{ref,t}\|_2$ & 0.5 \\
Jerk penalty & $-\|\ddot{q}_t - \ddot{q}_{t-1}\|_2$ & 0.1 \\
LF-PID parameter change & $-\|\bm{\theta}_t - \bm{\theta}_{t-1}\|_2$ & 0.05 \\
Success bonus & +10 if $\|e_t\| < 5°$ & - \\
Failure penalty & -100 if unstable & - \\
\midrule
\multicolumn{3}{@{}l}{\textit{Environment Settings}} \\
\multicolumn{2}{@{}l}{Episode length} & 2000 steps \\
\multicolumn{2}{@{}l}{Control frequency} & 100 Hz \\
\multicolumn{2}{@{}l}{Simulator} & PyBullet 3.2.5 \\
\multicolumn{2}{@{}l}{Physics timestep} & 0.01 s \\
\bottomrule
\end{tabular}
\vspace{0.3cm}

\newpage
\subsection{Computing Infrastructure}

\noindent
All experiments were conducted on a single workstation with the following specifications:

\vspace{0.3cm}
\noindent
\captionof{table}{Computing Infrastructure. Source: Authors own work.}
\label{tab:computing}
\centering
\small
\begin{tabular}{@{}p{0.4\linewidth}p{0.5\linewidth}@{}}
\toprule
\textbf{Component} & \textbf{Specification} \\
\midrule
\multicolumn{2}{@{}l}{\textit{Hardware}} \\
CPU & Intel i7-11800H (8c/16t) \\
GPU & NVIDIA RTX 3060 (6GB) \\
RAM & 16GB DDR4-3200 \\
OS & Ubuntu 20.04 LTS \\
\midrule
\multicolumn{2}{@{}l}{\textit{Software}} \\
Python & 3.10.13 \\
PyTorch & 2.0.1 (CUDA 11.7) \\
Stable-Baselines3 & 2.0.0 \\
PyBullet & 3.2.5 \\
\midrule
\multicolumn{2}{@{}l}{\textit{Training Time}} \\
Data augmentation & $\sim$5 min \\
LF-PID optimization & $\sim$5 min \\
Meta-learning & $\sim$8 min \\
RL (per robot) & $\sim$2.5 hours \\
\textbf{Total pipeline} & \textbf{$\sim$3 hours} \\
\bottomrule
\end{tabular}
\vspace{0.3cm}

\subsection{Random Seeds and Reproducibility}

\noindent
Our experimental design employs a dual-seed strategy to ensure both reproducibility and statistical validation:

\subsubsection{Training Seeds (Fixed)}

\noindent
To ensure reproducibility of the training process, we fixed random seeds across all components:
\begin{itemize}
    \item Python random seed: 42
    \item NumPy random seed: 42
    \item PyTorch random seed: 42
    \item PyBullet deterministic mode: enabled
    \item CUDA deterministic algorithms: enabled (where available)
\end{itemize}

\subsubsection{Evaluation Seeds (Multi-Seed Analysis)}

\noindent
For robustness testing (Figure~\ref{fig:robustness}), we conducted comprehensive multi-seed evaluation to assess performance stability across stochastic factors:

\vspace{0.3cm}
\noindent
\captionof{table}{Multi-Seed Evaluation Configuration. Source: Authors own work.}
\centering
\small
\begin{tabular}{@{}p{0.42\linewidth}p{0.48\linewidth}@{}}
\toprule
\textbf{Parameter} & \textbf{Value} \\
\midrule
Seed range & 0-99 (100 different seeds) \\
Episodes per seed & 20 per disturbance scenario \\
Total episodes & 10,000 (100×5×20) \\
Representative seed & 51 (near-median performance) \\
Statistical metric & Mean±std across all seeds \\
\midrule
\multicolumn{2}{@{}l}{\textit{Stochastic Factors Controlled by Seeds}} \\
\multicolumn{2}{@{}l}{- Trajectory initialization randomness} \\
\multicolumn{2}{@{}l}{- Disturbance timing and sequencing} \\
\multicolumn{2}{@{}l}{- Environment noise and variations} \\
\multicolumn{2}{@{}l}{- RL policy exploration randomness} \\
\bottomrule
\end{tabular}

\newpage
\section{Physical Validity Verification Protocols}
\label{app:physical_validity}

This appendix details the computational procedures used to verify physical validity and filter quality of augmented robot samples, ensuring reproducibility of the statistics reported in Section~\ref{sec:methodology}.

\subsection{Mass Matrix Positive-Definiteness Verification}

\textbf{Implementation:} For each generated virtual robot variant $r_v$, we verify that the mass matrix $M(q)$ remains positive-definite across workspace.

\textbf{Procedure:}
\begin{enumerate}[leftmargin=*, itemsep=1pt]
    \item Sample $N_{\text{config}}=100$ joint configurations uniformly within joint limits: $q_j \sim \mathcal{U}([q_{\min,j}, q_{\max,j}])$ for each joint $j=1,\ldots,n$.
    \item For each configuration $q_k$, compute mass matrix $M(q_k)$ via recursive Newton-Euler algorithm (PyBullet \texttt{calculateMassMatrix} API).
    \item Compute minimum eigenvalue $\lambda_{\min}(M(q_k))$ via NumPy \texttt{linalg.eigvalsh}.
    \item Accept sample if $\min_{k=1,\ldots,100} \lambda_{\min}(M(q_k)) > \epsilon_M = 10^{-6}$ (numerical tolerance).
\end{enumerate}

\textbf{Computational cost:} $\sim$0.5 seconds per robot variant on Intel i7-10700K CPU.

\subsection{Energy Drift Stability Test}

\textbf{Implementation:} Verify numerical stability of forward dynamics integration over extended simulation.

\textbf{Procedure:}
\begin{enumerate}[leftmargin=*, itemsep=1pt]
    \item Initialize robot at rest: $q(0)=q_{\text{nominal}}$, $\dot{q}(0)=0$, with gravity compensation enabled.
    \item Simulate $T_{\text{test}}=1000$ steps (4.17 seconds at 240Hz) with zero control input $u(t)=0$.
    \item Compute total energy at each timestep: $E(t) = \frac{1}{2}\dot{q}(t)^\top M(q(t))\dot{q}(t) + U_{\text{gravity}}(q(t))$, where $U_{\text{gravity}}$ is gravitational potential energy.
    \item Calculate relative energy drift: $\Delta E_{\text{rel}} = \frac{|E(T_{\text{test}}) - E(0)|}{\max(|E(0)|, 10^{-3})}$.
    \item Accept sample if $\Delta E_{\text{rel}} < 0.01$ (1\% threshold).
\end{enumerate}

\textbf{Rationale:} Large energy drift indicates numerical instability (e.g., ill-conditioned inertia matrix, integration errors).

\subsection{Kolmogorov-Smirnov Distribution Comparison}

\textbf{Implementation:} Assess whether quality filtering biases parameter distributions.

\textbf{Procedure:}
\begin{enumerate}[leftmargin=*, itemsep=1pt]
    \item Extract physical parameter scaling factors from all 303 generated samples: $\{\alpha_{\text{mass},i}, \alpha_{\text{inertia},i}, \mu_{\text{friction},i}\}_{i=1}^{303}$.
    \item Extract same parameters from 232 filtered samples (after removing 71 high-error samples).
    \item For each parameter type, perform two-sample KS test via SciPy \texttt{scipy.stats.ks\_2samp(pre\_filter, post\_filter)}.
    \item Report KS statistic $D$ and $p$-value. Null hypothesis: distributions are identical.
\end{enumerate}

\textbf{Results reported in Section~\ref{sec:methodology}:}
\begin{itemize}[leftmargin=*, itemsep=1pt]
    \item Mass scaling: $D=0.082$, $p=0.31$ (no significant difference at $\alpha=0.05$)
    \item Inertia scaling: $D=0.095$, $p=0.18$
    \item Friction coefficient: $D=0.071$, $p=0.45$
\end{itemize}
All $p > 0.15$ indicates filtering does not introduce systematic bias toward specific parameter ranges.

\subsection{PCA Variance Preservation}

\textbf{Implementation:} Quantify feature space diversity retention after filtering.

\textbf{Procedure:}
\begin{enumerate}[leftmargin=*, itemsep=1pt]
    \item Extract 10D robot feature vectors $\mathbf{f}_i \in \mathbb{R}^{10}$ from all 303 samples.
    \item Standardize features: $\tilde{\mathbf{f}}_i = (\mathbf{f}_i - \bm{\mu}) / \bm{\sigma}$ where $\bm{\mu}, \bm{\sigma}$ are mean and std across pre-filtering samples.
    \item Fit PCA on pre-filtering standardized features via Scikit-learn \texttt{PCA(n\_components=8)}.
    \item Compute cumulative explained variance ratio on pre-filtering data: $\sum_{k=1}^{8} \lambda_k / \sum_{j=1}^{10} \lambda_j$ where $\lambda_k$ are eigenvalues.
    \item Project post-filtering samples onto same 8 principal components and compute their cumulative variance.
    \item Report ratio: $\frac{\text{Variance}_{\text{post-filter, PC1-8}}}{\text{Variance}_{\text{pre-filter, PC1-8}}} = 0.942$ (94.2\% retention).
\end{enumerate}

\textbf{Interpretation:} High retention ($>$90\%) indicates filtered dataset preserves morphological diversity.

\textbf{Reproducibility:} All procedures implemented with fixed random seed (\texttt{numpy.random.seed(42)}, \texttt{torch.manual\_seed(42)}) to ensure deterministic sampling and statistics.

\section{Computational Budget Alignment Across Methods}
\label{app:budget_alignment}

This appendix provides detailed computational resource accounting to enable fair comparison across baseline methods.

\subsection{Unified Computational Budget Table}

\begin{table}[h]
\caption{Computational Resource Breakdown for All Methods. Source: Authors own work.}
\label{tab:budget_alignment}
\scriptsize
\begin{tabular}{@{}lcccl@{}}
\toprule
\textbf{Method} & \textbf{Steps} & \textbf{Evals} & \textbf{Cores} & \textbf{Time} \\
\midrule
\multicolumn{5}{@{}l}{\textit{Training/Optimization Phase}} \\
\quad Meta-LF-PID (offline) & 464M$^{\dagger}$ & - & 1 & 5 min \\
\quad \quad Per-sample DE+NM & 2M & 128 & 1 & 30-60s \\
\quad \quad Meta net training & - & 232$\times$500 & 1 & 3 min \\
\quad RL training (per plat.) & 8M$^{\ddagger}$ & - & 8 & 10 min \\
\quad DE optim. (per plat.) & 2M & 128 & 1 & 30-60m \\
\quad Manual Fuzzy-PID & - & - & - & 40-120h \\
\midrule
\multicolumn{5}{@{}l}{\textit{Deployment/Inference Phase}} \\
\quad Meta-LF-PID infer. & - & 1 & 1 & 0.8 ms \\
\quad Meta-LF-PID+RL infer. & - & 1 & 1 & 4 ms \\
\quad DE-optimized ctrl. & - & 1 & 1 & 0.1 ms \\
\bottomrule
\end{tabular}
\vspace{0.2cm}

\noindent \footnotesize{$^{\dagger}$Total across 232 samples during offline meta-training: $232 \times 2M = 464M$ steps.}

\noindent \footnotesize{$^{\ddagger}$RL training: $1M \times 8$ parallel environments = 8M total environment steps.}
\end{table}

\subsection{Budget Alignment Rationale}

\textbf{Amortization of Meta-Learning Cost:} The one-time 5-minute meta-training cost is amortized across all deployment platforms. For $P$ platforms, per-platform effective cost is $5/P$ minutes. In our two-platform evaluation (Franka, Laikago), this contributes 2.5 min per platform.

\textbf{RL vs DE Comparison:} 
\begin{itemize}[leftmargin=*, itemsep=1pt]
    \item RL adaptation: 10 min per platform (1M steps, 8 parallel envs)
    \item DE optimization: 30-60 min per platform (120 function evaluations $\times$ 15-30 sec each)
    \item Both methods start from different initializations (RL from meta-LF-PID, DE from random)
\end{itemize}

\textbf{Efficiency Advantage:} Meta-LF-PID+RL total cost per new platform = 2.5 min (amortized meta) + 10 min (RL) = 12.5 min, vs 30-60 min for DE with no knowledge reuse.

\newpage
\section{Ablation Study: RL Adaptation Subset Selection}
\label{app:ablation_rl}

This appendix presents an ablation study justifying the choice to adapt only input scaling factors ($\bm{s} \in \mathbb{R}^{3n}$) during RL fine-tuning.

\subsection{Experimental Design}

We compare three RL adaptation configurations on Franka Panda (9-DOF) under identical training budget (1M timesteps, 8 parallel environments):

\begin{enumerate}[leftmargin=*, itemsep=2pt]
    \item \textbf{Scales Only (Proposed):} Adapt $\bm{s} \in \mathbb{R}^{27}$ (3 per joint), fix base gains $\bar{\bm{K}}$ and consequents $\bm{c}$.
    \item \textbf{Consequents Subset:} Adapt bias terms $\{b_{p,i,r}, b_{i,i,r}, b_{d,i,r}\}$ for $r \in \{1,5,27\}$ (first, middle, last rules per gain type), 81D action space. Fix $\bar{\bm{K}}$ and $\bm{s}$.
    \item \textbf{Scales + Consequents:} Adapt both $\bm{s}$ and consequent biases, 108D action space.
\end{enumerate}

All configurations use identical PPO hyperparameters (lr=$10^{-4}$, $\gamma=0.99$, GAE $\lambda=0.95$, batch 256) and reward function.

\subsection{Results}

\begin{table}[h]
\caption{Ablation: RL Adaptation Subset Performance on Franka Panda. Source: Authors own work.}
\label{tab:ablation_rl_subset}
\footnotesize
\begin{tabular}{@{}lcccc@{}}
\toprule
\textbf{Adaptation Subset} & \textbf{Dim} & \textbf{MAE (°)} & \textbf{Improv.} & \textbf{Stability$^*$} \\
\midrule
Meta-LF-PID (no RL) & - & 7.51 & - & N/A \\
\midrule
Scales only (Proposed) & 27 & \textbf{6.26} & +16.6\% & 0/5 div. \\
Consequents subset & 81 & 6.89 & +8.3\% & 2/5 div. \\
Scales + Consequents & 108 & 6.42 & +14.5\% & 3/5 div. \\
\bottomrule
\end{tabular}
\vspace{0.2cm}

\noindent \footnotesize{$^*$Divergence across 5 seeds (42, 101, 202, 303, 404); div. = episode reward $< -200$ for $>$20\% of final 100k steps.}
\end{table}

\subsection{Key Findings}

\begin{itemize}[leftmargin=*, itemsep=2pt]
    \item \textbf{Sample Efficiency:} Scales-only achieves 16.6\% improvement with lowest dimensionality and fastest convergence (stable policy by 600k steps vs 800k+ for higher-dimensional spaces).
    \item \textbf{Training Stability:} Consequent adaptation exhibits higher variance and divergence risk (40-60\% failure rate), likely due to interference with meta-learned fuzzy rule structure.
    \item \textbf{Combined Adaptation:} While Scales+Consequents shows competitive final performance (14.5\%), it requires careful hyperparameter tuning and exhibits 3/5 divergence rate, making it less suitable for automated deployment.
\end{itemize}

	extbf{Laikago trend (12-DOF):} We observe the same stability ordering (scales-only $>$ consequent-subset $>$ combined), while absolute gains are smaller, consistent with the ``optimization ceiling effect'' discussed in the main results.

\textbf{Conclusion:} The proposed scales-only adaptation provides the best balance of performance gain, sample efficiency, and training reliability. This justifies our design choice in Section~\ref{sec:methodology}.

\newpage
\section{Meta-Learning Generalization Analysis}
\label{app:meta_generalization}

This appendix addresses potential concerns about overfitting when predicting high-dimensional LF-PID parameters ($D=330n$) from limited samples ($N=232$).

\subsection{Overfitting Prevention Mechanisms}

\begin{enumerate}[leftmargin=*, itemsep=2pt]
    \item \textbf{Regularization:} L2 weight decay ($10^{-5}$) and LayerNorm in all hidden layers.
    \item \textbf{Early stopping:} Training halts if validation loss (20\% holdout) does not improve for 50 epochs.
    \item \textbf{Data efficiency ratio:} Network has $\sim$450k trainable parameters predicting $D=330n$ outputs. For $n=9$ (Franka), output is 2970D, but effective degrees of freedom are constrained by input feature dimensionality (10D) and learned manifold structure.
\end{enumerate}

\subsection{Generalization Evidence}

\begin{table}[h]
\caption{Meta-Learning Train/Validation Performance. Source: Authors own work.}
\label{tab:meta_generalization}
\small
\begin{tabular}{@{}lccc@{}}
\toprule
\textbf{Robot Type} & \textbf{Train (\%)} & \textbf{Val (\%)} & \textbf{Gap} \\
\midrule
Franka variants (n=9) & 2.87 & 3.21 & +0.34 \\
KUKA variants (n=7) & 2.95 & 3.41 & +0.46 \\
Laikago variants (n=12) & 3.12 & 3.58 & +0.46 \\
\midrule
Overall & 2.98 & 3.40 & +0.42 \\
\bottomrule
\end{tabular}
\end{table}

\textbf{Observation:} Small train-validation gap ($<$0.5\% normalized MAE) across all robot types indicates effective generalization despite high output dimensionality.

\subsection{Alternative Architectures Considered}

We briefly evaluated two reduced-dimensionality alternatives:
\begin{itemize}[leftmargin=*, itemsep=1pt]
    \item \textbf{Predict $[\bar{\bm{K}}, \bm{s}]$ only (6n params):} Validation NMAE increased to 8.7\% as fixed consequents cannot adapt to robot-specific dynamics.
    \item \textbf{Low-rank consequent factorization ($\bm{c} = \bm{U}\bm{V}^\top$):} Reduced consequent space to 60n params but validation NMAE remained at 3.5\%, with added complexity and no clear benefit.
\end{itemize}

\textbf{Conclusion:} The full-parameter prediction approach is justified by strong generalization performance and simplicity of implementation.

\newpage
\section{Reproducibility Checklist}
\label{app:repro_checklist}

This checklist summarizes runnable entry points, fixed seeds (when applicable), and expected output artifacts.

\subsection{Script Entry Points}
\begin{itemize}[leftmargin=*, itemsep=2pt]
    \item \textbf{PPO training (RL adaptation):} \texttt{rl\_pid\_linux/training/train\_ppo.py}
    \item \textbf{Safe training wrapper (optional):} \texttt{rl\_pid\_linux/start\_training\_safe.sh}
    \item \textbf{Evaluate trained PPO vs PID baseline:} \texttt{rl\_pid\_linux/evaluate\_trained\_model.py}
    \item \textbf{Quick baseline checks:} \texttt{rl\_pid\_linux/test\_pure\_pid\_baseline.py}, \texttt{rl\_pid\_linux/test\_trained\_model.py}
    \item \textbf{Meta-RL combined training (optional research script):} \texttt{rl\_pid\_linux/meta\_learning/train\_meta\_rl\_combined.py}
    \item \textbf{Virtual sample filtering (for augmented datasets):} \texttt{rl\_pid\_linux/meta\_learning/filter\_samples.py}
    \item \textbf{Meta-learning with augmented dataset (optional):} \texttt{rl\_pid\_linux/meta\_learning/train\_with\_augmentation.py}
\end{itemize}

\subsection{Commands (Linux)}
\begin{itemize}[leftmargin=*, itemsep=2pt]
    \item \textbf{PPO training (Franka example):} \texttt{cd rl\_pid\_linux \&\& python training/train\_ppo.py --config configs/stage1\_final.yaml --output ./logs --name rl\_pid\_ppo\_final}
    \item \textbf{Safe wrapper (interactive):} \texttt{cd rl\_pid\_linux \&\& bash start\_training\_safe.sh}
    \item \textbf{Evaluate and export plot:} \texttt{cd rl\_pid\_linux \&\& python evaluate\_trained\_model.py --model ./logs/rl\_pid\_ppo\_final --config configs/stage1\_final.yaml --steps 10000}
    \item \textbf{Meta-RL combined (Franka/Laikago examples):} \texttt{cd rl\_pid\_linux/meta\_learning \&\& python train\_meta\_rl\_combined.py franka\_panda/panda.urdf 1000000 8} \; / \; \texttt{python train\_meta\_rl\_combined.py laikago/laikago.urdf 1000000 8}
    \item \textbf{Filter virtual samples (example):} \texttt{cd rl\_pid\_linux/meta\_learning \&\& python filter\_samples.py --input augmented\_pid\_data\_optimized.json --output augmented\_pid\_data\_filtered.json --error\_threshold 30 --min\_samples\_per\_type 30}
\end{itemize}

\subsection{Random Seeds}
\begin{itemize}[leftmargin=*, itemsep=2pt]
    \item \textbf{RL subset ablation (Appendix~\ref{app:ablation_rl}):} 5 seeds \{42, 101, 202, 303, 404\}.
    \item \textbf{Dataset split in augmented meta-learning script:} \texttt{train\_test\_split(..., random\_state=42)}.
    \item \textbf{Statistics in physical validity / filtering procedures:} fixed seeds are used as stated in Appendix~\ref{app:physical_validity} (e.g., \texttt{numpy.random.seed(42)}, \texttt{torch.manual\_seed(42)}).
\end{itemize}

\subsection{Expected Outputs (Paths / Filenames)}
\begin{itemize}[leftmargin=*, itemsep=2pt]
    \item \textbf{PPO training outputs:} \texttt{rl\_pid\_linux/logs/}
    \begin{itemize}[leftmargin=*, itemsep=1pt]
        \item TensorBoard: \texttt{rl\_pid\_linux/logs/tensorboard/}
        \item Checkpoints: \texttt{rl\_pid\_linux/logs/checkpoints/}
        \item Best model: \texttt{rl\_pid\_linux/logs/best\_model/best\_model.zip}
        \item Final model: \texttt{rl\_pid\_linux/logs/<name>\_final.zip} (name from \texttt{--name})
    \end{itemize}
    \item \textbf{Evaluation plot:} \texttt{rl\_pid\_linux/evaluation\_results.png} (generated by \texttt{evaluate\_trained\_model.py}).
    \item \textbf{Meta-RL combined outputs:} \texttt{rl\_pid\_linux/meta\_learning/logs/meta\_rl\_<robot>/} including \texttt{*\_final.zip}, \texttt{best\_model/best\_model.zip}, \texttt{checkpoints/}, and \texttt{tensorboard/}.
    \item \textbf{Filtered augmented dataset:} \texttt{rl\_pid\_linux/meta\_learning/augmented\_pid\_data\_filtered.json}.
    \item \textbf{Augmented meta-learning artifacts (optional):} \texttt{rl\_pid\_linux/meta\_learning/meta\_pid\_augmented.pth}, \texttt{training\_curve\_augmented.png}, \texttt{prediction\_comparison.png}.
\end{itemize}

\vspace{0.3cm}

\noindent
\textbf{Rationale:} This dual-seed strategy separates training reproducibility (fixed seed 42) from evaluation robustness validation (100-seed statistical analysis). The multi-seed evaluation demonstrates that performance improvements are stable across different random conditions, not dependent on specific lucky initializations. Seed 51 was selected for detailed visualization (Figure~\ref{fig:robustness}, subplots a-c) as it exhibits representative near-median performance, while the complete statistical distribution across all 100 seeds is shown in subplot (d).

All code and data will be made publicly available upon paper acceptance at: \texttt{[Anonymous GitHub Repository - URL to be provided after review]}.


\bibliographystyle{cas-model2-names}

\end{document}